\documentclass{article}

\usepackage{arxiv}

\usepackage[utf8]{inputenc} 
\usepackage[T1]{fontenc}    
\usepackage{hyperref}       
\usepackage{url}            
\usepackage{booktabs}       
\usepackage{amsfonts}       
\usepackage{nicefrac}       
\usepackage{microtype}      
\usepackage{lipsum}		
\usepackage{graphicx}
\usepackage{natbib}
\usepackage{doi}

\usepackage{dcolumn}
\usepackage{lineno,hyperref}
\usepackage{colortbl}
\usepackage{color}
\usepackage[table,xcdraw]{xcolor}
\usepackage{microtype}
\usepackage{multirow}
\usepackage{rotating}
\usepackage{booktabs}
\usepackage{pdfpages}
\usepackage{lipsum}
\usepackage{graphicx}
\usepackage{array}
\usepackage{booktabs}
\usepackage{ltablex}
\usepackage{longtable}
\usepackage{float}
\usepackage{listings}
\usepackage[colorinlistoftodos]{todonotes}

\usepackage{tikz}
\usetikzlibrary{shapes,arrows}
\usepackage{multicol}
\usepackage{placeins}
 \usepackage{algorithm}
\usepackage{algorithmic}
\usepackage{subcaption}

\newcolumntype{d}[1]{D{.}{.}{#1}}

\title{Label Noise Detection under the Noise at Random Model with Ensemble Filters}

\author{
{
\hspace{1mm}Kecia G. Moura, 
\hspace{1mm}Ricardo B. C. Prudêncio, 
\hspace{1mm}George D. C. Cavalcanti}
\\
	Centro de Informática, Universidade Federal de Pernambuco, CIn-UFPE, Recife, PE, Brazil \\
    \texttt{  \string{kgm, rbcp, gdcc\string}@cin.ufpe.br} \\
}

\renewcommand{\shorttitle}{Label noise detection  under  the Noise at Random model with ensemble filters}

\hypersetup{
pdftitle={Label noise detection under the NAR model with ensemble filters},
pdfsubject={cs.LG},
pdfauthor={Kecia G.~Moura,  Ricardo B. C.~Prudêncio,  George D. C.~Cavalcanti},
pdfkeywords={Label noise, Noise detection, Ensemble methods, Noise at Random, Ensemble noise filtering},
}

\begin{document}
\maketitle

\begin{abstract}
Label noise detection has been widely studied in Machine Learning because of its importance  in improving training data quality. Satisfactory noise detection has been achieved by adopting ensembles of classifiers. In this approach, an instance is assigned as mislabeled if a high proportion of members in the pool misclassifies it. Previous authors have empirically evaluated this approach; nevertheless, they mostly assumed that label noise is generated completely at random in a dataset. This is a strong assumption since other types of label noise are feasible in practice and can influence noise detection results. This work investigates the performance of ensemble noise detection under two different noise models: the Noisy at Random (NAR), in which the probability of label noise depends on the instance class, in comparison to the Noisy Completely at Random model, in which the probability of label noise is entirely independent. In this setting, we investigate the effect of class distribution on noise detection performance since it changes the total noise level observed in a dataset under the NAR assumption. Further, an evaluation of the ensemble vote threshold is conducted to contrast with the most common approaches in the literature. In many performed experiments, choosing a noise generation model over another can lead to different results when considering aspects such as class imbalance and noise level ratio among different classes.
\end{abstract}


\section{Introduction}
\label{sec:introduction}
Data quality is of great importance for {ML} applications and, in particular, for classification tasks. Conventionally in these tasks, a training set of labeled instances is given as input to an {ML} algorithm, which will acquire useful knowledge to make predictions for new instances. In practice, real-world datasets frequently contain irregularities such as incompleteness, noise, and data inconsistencies that impact {ML} performance \citep{Han}. In this light, noise detection and filtering are quite relevant techniques for {ML} \citep{Zhu-attr}. 

According to the literature, noise may occur in both attributes and classes \citep{Zhu-attr}. This work focuses on the latter problem, in which an unknown proportion of instances in a dataset are mislabeled because of different reasons. This is a relevant problem since label noise can harm the identification of  true class boundaries in a problem, increase the chance of overfitting, and affect learning performance in general \citep{survey}.  

Previous works successfully adopted the  \textit{classification noise filtering} method \citep{Brodley}\citep{Sluban-diversity}\citep{unlabled2018} for label noise detection, widespread in the literature. In this approach, mislabeled instances in a dataset are identified according to the output results of a classifier or an ensemble of classifiers. For example, in the majority vote for ensemble noise detection, an instance is marked as mislabeled if most classifiers in a pool incorrectly classify it. In the consensus vote for ensemble noise detection, a record is considered noisy if all classifiers in the pool misclassify it.

As with most ML tasks, empirical evaluation is also crucial in the context of noise detection techniques. Producing a ground-truth dataset for evaluation usually requires additional domain experts to decide which instances were mislabeled. This process can be costly, and experts are not always available. This problem is mitigated when artificial datasets are used or when simulated noise is injected into a dataset in a controlled way. The investigation of how noise influences the learning process is simplified when a systematic addition of noise is performed \citep{GARCIA2019}. 
 
Label noise can be injected into a dataset by assuming three distinct models of noise \cite{survey}:   
(i) Noisy Completely at Random (NCAR), in which the probability of an instance being noisy is random, (ii) Noisy at Random (NAR), the probability of an instance being noisy depends on its label, and (iii) Non-Noisy at Random (NNAR), the probability of an instance being noisy also depends on its attributes. In many previous works \citep{Sluban-diversity} \citep{Brodley} \citep{saesSMOTE} \citep{GARCIA2019}, a single noise model is chosen over another to perform experiments. Nevertheless, it is usually not clear how this choice can affect experimental results. Additionally, other aspects, like class distribution,  can impact the distribution of noise differently in a dataset depending on the noise type considered. For instance, a human supervisor may find it more difficult to label records from the minority class than the majority class in some contexts.

In this work, it is investigated how noise models can influence noise detection experiments under different aspects. In contrast to previous studies, the influence of distinct label noise models on ensemble noise detection is evaluated in this research under various contexts such as class imbalance, noise distribution, ensemble thresholds, and percentage of noise in data. 
It is shown that different results are achieved depending on the context. For instance, even under the same noise model (e.g., NAR), a detection technique may have quite distinct performance results if class imbalance changes (e.g., NAR with imbalanced vs. NAR with balanced class distributions). 

The remainder of this paper is organized as follows. In Section~\ref{sec:background}, an overview of label noise detection is presented. 
The proposed methodology is described in  Section~\ref{sec:experiments}. 
Experiments are presented in Section~\ref{sec:results}. 
Finally, Section~\ref{sec:conclusion} summarizes the paper and presents future work.

\section{Related Works}
\label{sec:background}

In \cite{Zhu-attr}, two types of noise are distinguished for supervised learning datasets: attribute (or feature) and class (or label) noise. The former is present in one or more features due to absent, incorrect, or missing values. In turn, label noise can be generated because of many reasons, such as the low reliability of human experts during labeling, incomplete information, communication problems, among others \citep{Sluban-advances}. The presence of noise in the training dataset can lead to an increase in processing time, higher model complexity, and the chance of overfitting, which will then deteriorate the predictive performance \citep{LORENA2004}.

According to the literature, removing examples with feature noise is not as beneficial as label noise detection. This occurs since the values of non-noisy features can be helpful in the classification process and because there is only one label, while there are many attributes \citep{survey}. Besides, feature noise can later give rise to label noise. Hence, this work will concentrate on the label noise problems. From now on, label noise is also referred to as noise.

This section initially presents a brief review of standard label noise detection techniques in the literature (Section~\ref{sec:background_detection_techniques}). Previous works have commonly evaluated such methods by performing experiments in which noise is intentionally injected into a dataset. A model is required in these experiments to control how the label noise is distributed across the dataset instances. Different label noise models are presented in Section~\ref{sec:background_noise_taxonomy}. Finally, the scope of this paper and its contributions are presented in Section~\ref{sec:main_contribuitions}. 

\subsection{Noise Detection Approaches}
\label{sec:background_detection_techniques}

Several techniques have already been developed for dealing with label noise. According to \cite{survey}, two broad strategies can be used: (1) the algorithm-level strategy, i.e., designing classifiers that are more robust and noise-tolerant, and (2) the data-level strategy, i.e., performing data cleaning by filtering noisy instances as a preprocessing step.

\subsubsection{Algorithm-level Approach}

Some learning algorithms are naturally more tolerant to noise, which can be used as a benefit in the presence of label noise. Ensemble methods like \textit{bagging} have the diversity increased when noise is present, which helps to cope with mislabeled examples. The decision tree pruning process is also more robust to noisy data as it has been shown that this technique decreases the influence of label noise since it prevents data overfitting \citep{abellan-baggingDT}.

Even more robust learning algorithms can be derived by including the noise information during the learning process. In \cite{B2016star}, for example, it is proposed the \textit{generalized robust Logistic Regression} (gLR), in which the exponential  distribution was adopted to model noise in such a way that points closer to the decision boundary have a relatively higher chance of being mislabeled. In the \textit{robust Kernel Fisher Discriminant} \citep{FisherLawrence:2001}, a probability of the label being noisy is derived by applying an Expectation-Maximization algorithm. In the  \textit{robust kernel logistic regression} \citep{BOOTKRAJANG2014}, the optimal hyperparameters of the method are automatically determined using Multiple Kernel Learning and Bayesian regularization techniques. In \cite{B2016star}, a logistic regression classifier is built by employing a noise model based on a mixture of Gaussians. In \cite{biggio2011}, the \textit{Label Noise robust SVMs} deal with noise present in data by correcting the kernel matrix with a specially structured matrix based on the information regarding the level of noise in the dataset. 

The approaches above directly model label noise during the learning process. Although the advantage of those methods is to use prior knowledge regarding a noise model and its consequences \citep{survey}, it increases the complexity of learning algorithms and can lead to overfitting, because of the additional parameters of the training data model.

\subsubsection{Data-level Approach}

While the algorithm-level strategy aims to implement robust models using some available information related to the noise present in data, the data-level approach handles noisy data before the training process. The algorithm-level methods are less versatile, as not all algorithms have robust versions. In turn, the data-level strategy has the advantage of considering the noise filtering and the learning phase as distinct steps. Hence, it avoids using polluted instances during the learning process, improving both predictive performance and computational cost. Also, filtering approaches are usually cheap and easy to implement \citep{survey}. Thus, our work will be focused on the data-level category.

Label noise filtering can be performed in a variety of ways. For example: by using complexity measures for the records \citep{sun-adHocMeasure} \citep{Smith2014}; partitioning approaches for removing mislabeled instances in large datasets \citep{zhu-large}\citep{GARCIAGIL2019bigData}; filtering noisy examples by verifying the impact of the removal on the learning process \citep{LOOPC}; using neighborhood-based algorithms to remove instances that are distant from the ones of the same class \citep{Wilson2000,Kanj2016ENN}, among others.

In our work, we adopted the ensemble approach for noise filtering, in which instances are removed when a certain number of algorithms misclassifies them \citep{YUAN2018psma}. This approach has been widely chosen \citep{Brodley}\citep{zhu-large}\citep{comitee} \citep{Sluban-advances}\citep{Sluban2014}\citep{saesSMOTE}\citep{YUAN2018psma}\citep{unlabled2018}, as it overcomes the problem of relying on a single classifier for noise filtering. Using only one classifier for noise filtering can cause the removal of too many instances. The ensemble approach improves noise detection since an instance is likely to have been incorrectly labeled if distinct classifiers disagree on their predictions. The ensemble noise filtering applies the $k$-fold cross-validation, i.e., in $k$ repetitions, $k-1$ folds of the dataset are used for training each algorithm in the ensemble, and the remaining fold is used for validation. Then, all records are classified by all algorithms in the pool, and the observed errors are taking into account to remove instances. 

An essential issue in ensemble-based noise filtering is how many misclassifications are assumed to consider an instance as noisy. There are two common choices in the literature: the \textit{consensus} and the \textit{majority} vote \citep{unlabled2018}\citep{Sluban-diversity}\citep{YUAN2018psma}. Whereas the majority vote classifies an instance as incorrectly labeled if a majority of the algorithms in the pool misclassifies it, the consensus vote requires that all classifiers have misclassified the record. These vote techniques can produce different results. As the consensus requires a higher agreement of classifiers, it tends to remove a few instances. On the other hand, the majority vote may throw out too many instances, including noise-free ones that could be relevant.

The trade-off between choosing the majority or the consensus approach can be replaced by the problem of selecting a vote threshold. The majority and the consensus vote are special cases, when thresholds are 50\% and 100\%, respectively. The adequate vote threshold would be related to the expected proportion of noisy instances in a dataset. Nevertheless, few works have investigated the influence of varying the vote thresholds. For example, in \cite{threshold2005} and \cite{threshold2018}, the authors showed that selecting appropriate values of the vote threshold usually performed better than using the standard filtering approaches. 

\subsection{Noise Models}
\label{sec:background_noise_taxonomy}

In real-world applications, evaluating whether an example is noisy or not generally requires the examination of domain specialists. Nonetheless, this is not always feasible as they may not be available. Moreover, consulting a specialist tends to increase the duration and cost of the preprocessing step. This problem is mitigated when artificial datasets are used, or simulated noise is injected into a dataset in a controlled way. The study and further validation of noise detection techniques and noise models' influence on the learning process are simplified when a systematic addition of noise is performed. 

In order to do so, it is imperative to choose the method by which the noise will be inserted into a dataset.

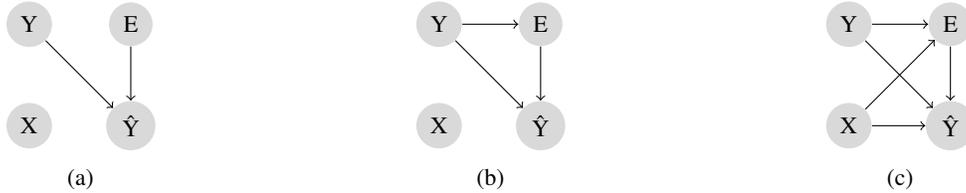
\begin{figure}[htp]
\begin{subfigure}{0.33\linewidth}
\centering
\begin{tikzpicture}[scale=.9, transform shape]
\tikzstyle{every node} = [circle, fill=gray!30]
\node (x) at (0, 0) {X};
\node (y1) at (1.5, 0) {\^Y};
\node (y) at (0, 1.5) {Y};
\node (e) at (1.5,1.5) {E};
\foreach \from/\to in {y/y1, e/y1}
\draw [->] (\from) -- (\to);
\end{tikzpicture}
\caption{}
\label{fig:survey_a}
\end{subfigure}%
\begin{subfigure}{0.33\linewidth}
\centering
\begin{tikzpicture}[scale=.9, transform shape]
\tikzstyle{every node} = [circle, fill=gray!30]
\node (x) at (0, 0) {X};
\node (y1) at (1.5, 0) {\^Y};
\node (y) at (0, 1.5) {Y};
\node (e) at (1.5,1.5) {E};
\foreach \from/\to in {y/y1, y/e, e/y1}
\draw [->] (\from) -- (\to);
\end{tikzpicture}\caption{}
\label{fig:survey_b}
\end{subfigure}
\begin{subfigure}{0.33\linewidth}
\centering
\begin{tikzpicture}[scale=.9, transform shape]
\tikzstyle{every node} = [circle, fill=gray!30]
\node (x) at (0, 0) {X};
\node (y1) at (1.5, 0) {\^Y};
\node (y) at (0, 1.5) {Y};
\node (e) at (1.5,1.5) {E};
\foreach \from/\to in {y/y1, y/e,  x/e, x/y1, e/y1}
\draw [->] (\from) -- (\to);
\end{tikzpicture}\caption{}
\label{fig:survey_c}
\end{subfigure}
\caption{Statistical taxonomy of label noise according to \cite{survey}. (a)  NCAR,  (b)  NAR,  and  (c)  NNAR. \textit{X} denotes the vector of features, \textit{Y} is the true class, \textit{\^Y} is the observed label, and \textit{E} is a binary variable telling whether a labeling error occurred.   Arrows  report  statistical dependencies. }
\label{fig:survey_p}
\end{figure}

In \cite{survey}, the authors provided a taxonomy of label noise models, reflecting the distribution of noisy instances in a dataset. The three models are shown in Figure \ref{fig:survey_p}. Let \textit{X} be the vector of features, \textit{Y} the true class, \textit{\^Y} the observed label, and \textit{E} a binary variable indicating if a labeling error occurred.  Each model has a different assumption on how noise is generated.

\begin{enumerate}
\item Noisy Completely at Random ({NCAR}): the occurrence of a mislabeled instance is independent of the instance's attributes and class. Mislabeled records are uniformly present  across the instance space. In a binary classification problem, for example, there will exist the same proportion of mislabeled instances in both classes. In other words, as shown in Figure  \ref{fig:survey_a}, the occurrence of an error \textit{E} is independent
of the other random variables, including the true class itself (\textit{Y}). For this model, the mislabeled instance probability is given by $p_{e} = P(E = 1) = P(Y  \neq  \textit{\^Y})$.

\item Noisy at Random ({NAR}): the labeling errors probability depends on the instance class, although it is not dependent on records' attributes. Once mislabeling is conditional to instance classes, it allows us to model asymmetric label noise, i.e., when samples from certain classes are more prone to be mislabeled. This model could be applied, for example, to simulate mislabeling classification that is often verified in medical case-control studies where the misclassification of disease outcome may be unrelated to risk factor exposure (non-differential) \citep{differencial}. As shown in Figure \ref{fig:survey_b}, \textit{E} is still independent of \textit{X} but it is conditioned by \textit{Y}. For this model, the mislabeled instance probability is given by $p_{e} = P(E = 1) = \sum_{y \in Y}   P(Y = y)P(E = 1|Y = y)$.

\item Noisy not at Random ({NNAR}): the probability of an error occurrence depends not only on the instance class but also on the instance attributes. In this case, for example, samples are more likely to be mislabeled when they are similar to records of another class or when they are located in certain regions of the instance space. By applying this model, it is possible to simulate mislabeling near classification boundaries or in low-density regions. It also can be used for medical case-control studies where the misclassification of disease outcome may be related to risk factor exposure (differential) \citep{differencial}.
As can be seen in Figure \ref{fig:survey_c}, this is a more complex model, where \textit{E} depends  on  both \textit{X} and
\textit{Y}, i.e., labeling errors are more likely for certain classes and in certain regions  of  the \textit{X} space.

\end{enumerate}

It is usually quite challenging to identify the kind of noise present in a dataset without any background knowledge. Nevertheless, it is crucial to evaluate how sensitive noise detection techniques are to the noise distribution in a dataset. In this work, analyses regarding the {NAR} and {NCAR} models were performed in different contexts. NNAR scenarios are equally relevant, although more diverse in terms of assumptions that relate to the instances' attributes and the chance of label noise. This work focused on the {NAR} and {NCAR} model to provide controlled experimental scenarios and investigate pertinent aspects (e.g., class distribution, noise level per class) that can impact label noise detection. Once deeply studied, such contexts can be extended in future work to cover the NNAR assumption.

\subsection{Scope and Contribution}
\label{sec:main_contribuitions}

As detailed previously,  several works have extensively studied approaches to better handle noise detection either by developing noise-tolerant algorithms or by identifying and filtering data irregularities in a preprocessing step.
In most previous studies, artificial noise is randomly injected into data to evaluate proposed systems. 
This work delivers relevant findings on how different noise models affect noise detection, indicating that new noise handlers systems should be evaluated in a broader context. In opposite to many related studies, which usually focus on creating new detectors, this paper analyses label noise detection under NCAR and NAR models and their behavior in different settings such as class imbalanced data, amount of noise, and noise distribution per class.

The current work is focused on evaluating ensemble filtering techniques, regarding different aspects that can impact the noise detection performance, such as the noise model, the class imbalance ratio and the noise level per class. Previous related works are commonly limited to evaluate ensemble filtering techniques assuming the NCAR model, such as \cite{Sluban-diversity} and \cite{unlabled2018}. Differently, our work investigates the performance of noise filters under the NAR model, in which noise level can vary depending on the class. Additionally, we addressed other aspects like class imbalance ratio and noise level per class, to investigate the impact on noise detection performance, also depending on the noise model assumed in the experiments. 

Previous works on ensemble noise filtering are also limited to evaluate and selecting between the majority or the consensus approach. The trade-off between choosing the majority or the consensus approach can be replaced by the problem of choosing a vote threshold. The majority and the consensus are special cases, when thresholds are 50\% and 100\%, respectively. The adequate vote threshold would be related to the expected proportion of noisy instances in a dataset. Nevertheless, few works have investigated the influence of varying the vote thresholds. For instance, in \cite{threshold2005} and \cite{threshold2018}, the authors showed that selecting adequate values of the vote threshold usually performed better than using the standard filtering approaches. The choice of the vote threshold is another important aspect that is addressed in the current work.




As it will be seen, our work produced important findings which can be considered when developing new noise detection techniques, evaluating existing approaches, and modeling specific real-world problems.


\section{Proposed Methodology}
\label{sec:experiments}

This section details the experimental setup adopted in our work to evaluate the ensemble noise detectors under {NCAR} and {NAR} models. Different aspects are jointly considered in the performed evaluation, which were not properly evaluated yet in the literature: (1) the class imbalance ratio in a dataset; (2) the noise ratio comparing the majority and minority classes; (3) the noise model itself.

\begin{figure}[t]
\begin{center}
\tikzset{every picture/.style={line width=0.75pt}} 

\begin{tikzpicture}[x=0.75pt,y=0.75pt,yscale=-1,xscale=1]

\draw   (92.67,144.67) .. controls (92.67,140.25) and (96.25,136.67) .. (100.67,136.67) -- (154.67,136.67) .. controls (159.08,136.67) and (162.67,140.25) .. (162.67,144.67) -- (162.67,168.67) .. controls (162.67,173.08) and (159.08,176.67) .. (154.67,176.67) -- (100.67,176.67) .. controls (96.25,176.67) and (92.67,173.08) .. (92.67,168.67) -- cycle ;
\draw   (183.33,146.6) .. controls (183.33,142.4) and (186.74,139) .. (190.93,139) -- (290.4,139) .. controls (294.6,139) and (298,142.4) .. (298,146.6) -- (298,169.4) .. controls (298,173.6) and (294.6,177) .. (290.4,177) -- (190.93,177) .. controls (186.74,177) and (183.33,173.6) .. (183.33,169.4) -- cycle ;
\draw   (269.33,80) .. controls (269.33,75.58) and (272.92,72) .. (277.33,72) -- (331.33,72) .. controls (335.75,72) and (339.33,75.58) .. (339.33,80) -- (339.33,104) .. controls (339.33,108.42) and (335.75,112) .. (331.33,112) -- (277.33,112) .. controls (272.92,112) and (269.33,108.42) .. (269.33,104) -- cycle ;
\draw   (383,128.67) .. controls (383,124.25) and (386.58,120.67) .. (391,120.67) -- (445,120.67) .. controls (449.42,120.67) and (453,124.25) .. (453,128.67) -- (453,152.67) .. controls (453,157.08) and (449.42,160.67) .. (445,160.67) -- (391,160.67) .. controls (386.58,160.67) and (383,157.08) .. (383,152.67) -- cycle ;
\draw   (381,26.67) .. controls (381,22.25) and (384.58,18.67) .. (389,18.67) -- (443,18.67) .. controls (447.42,18.67) and (451,22.25) .. (451,26.67) -- (451,50.67) .. controls (451,55.08) and (447.42,58.67) .. (443,58.67) -- (389,58.67) .. controls (384.58,58.67) and (381,55.08) .. (381,50.67) -- cycle ;
\draw   (471,26.67) .. controls (471,22.25) and (474.58,18.67) .. (479,18.67) -- (538,18.67) .. controls (542.42,18.67) and (546,22.25) .. (546,26.67) -- (546,50.67) .. controls (546,55.08) and (542.42,58.67) .. (538,58.67) -- (479,58.67) .. controls (474.58,58.67) and (471,55.08) .. (471,50.67) -- cycle ;
\draw   (473,83.67) .. controls (473,79.25) and (476.58,75.67) .. (481,75.67) -- (535,75.67) .. controls (539.42,75.67) and (543,79.25) .. (543,83.67) -- (543,107.67) .. controls (543,112.08) and (539.42,115.67) .. (535,115.67) -- (481,115.67) .. controls (476.58,115.67) and (473,112.08) .. (473,107.67) -- cycle ;
\draw   
(127,104) -- (127,134) ;
\draw [shift={(127,134)}, rotate = 270] [fill={rgb, 255:red, 0; green, 0; blue, 0 }  ][line width=0.75]  [draw opacity=0] (8.93,-4.29) -- (0,0) -- (8.93,4.29) -- cycle    ;



\draw    (162.67,156.67) -- (181.33,156.67) ;
\draw [shift={(183.33,156.67)}, rotate = 180] [fill={rgb, 255:red, 0; green, 0; blue, 0 }  ][line width=0.75]  [draw opacity=0] (8.93,-4.29) -- (0,0) -- (8.93,4.29) -- cycle    ;

\draw    (316,112) -- (316,141) -- (379.33,140.35) ;
\draw [shift={(381.33,140.33)}, rotate = 539.4200000000001] [fill={rgb, 255:red, 0; green, 0; blue, 0 }  ][line width=0.75]  [draw opacity=0] (8.93,-4.29) -- (0,0) -- (8.93,4.29) -- cycle    ;

\draw    (316,72) -- (316,39) -- (379,39) ;
\draw [shift={(381,39)}, rotate = 180] [fill={rgb, 255:red, 0; green, 0; blue, 0 }  ][line width=0.75]  [draw opacity=0] (8.93,-4.29) -- (0,0) -- (8.93,4.29) -- cycle    ;


\draw    (238,140) -- (238.33,97.33) -- (266,97.02) ;
\draw [shift={(268,97)}, rotate = 539.36] [fill={rgb, 255:red, 0; green, 0; blue, 0 }  ][line width=0.75]  [draw opacity=0] (8.93,-4.29) -- (0,0) -- (8.93,4.29) -- cycle    ;

\draw    (451.67,39) -- (470.33,39) ;
\draw [shift={(472.33,39)}, rotate = 180] [fill={rgb, 255:red, 0; green, 0; blue, 0 }  ][line width=0.75]  [draw opacity=0] (8.93,-4.29) -- (0,0) -- (8.93,4.29) -- cycle    ;

\draw   (473,144.67) .. controls (473,140.25) and (476.58,136.67) .. (481,136.67) -- (535,136.67) .. controls (539.42,136.67) and (543,140.25) .. (543,144.67) -- (543,168.67) .. controls (543,173.08) and (539.42,176.67) .. (535,176.67) -- (481,176.67) .. controls (476.58,176.67) and (473,173.08) .. (473,168.67) -- cycle ;
\draw    (419,120) -- (477.6,60.1) ;
\draw [shift={(479,58.67)}, rotate = 494.37] [fill={rgb, 255:red, 0; green, 0; blue, 0 }  ][line width=0.75]  [draw opacity=0] (8.93,-4.29) -- (0,0) -- (8.93,4.29) -- cycle    ;

\draw    (509,59) -- (509,73) ;
\draw [shift={(509,75)}, rotate = 270] [fill={rgb, 255:red, 0; green, 0; blue, 0 }  ][line width=0.75]  [draw opacity=0] (8.93,-4.29) -- (0,0) -- (8.93,4.29) -- cycle    ;

\draw    (507.33,115.67) -- (507.33,134.67) ;
\draw [shift={(507.33,136.67)}, rotate = 270] [fill={rgb, 255:red, 0; green, 0; blue, 0 }  ][line width=0.75]  [draw opacity=0] (8.93,-4.29) -- (0,0) -- (8.93,4.29) -- cycle    ;

\draw (127,92) node  [align=left] {Data};
\draw (240.67,156) node [scale=0.9] [align=left] { \ \ \ Generate IR \\by undersampling};
\draw (127.67,162.67) node [scale=0.9] [align=left] {Clean \\ data \\};
\draw (304.33,92) node  [align=left] {Split data};
\draw (341,153) node  [align=left] {Testing};
\draw (342,28) node  [align=left] {Training};
\draw (418.33,37.67) node [scale=0.9] [align=left] { \ \ \ Train \\algorithms };
\draw (418,140.67) node [scale=0.9] [align=left] {Inject \\noise };
\draw (506.5,38.67) node [scale=0.9] [align=left] { Select vote\\ \ threshold };
\draw (508,95.67) node [scale=0.9] [align=left] {Ensemble \\prediction};
\draw (509,155.67) node [scale=0.9] [align=left] {Evaluation};

\end{tikzpicture}
\caption{General experimental protocol.\label{fig:graph-protocol}}
\end{center}
\end{figure}
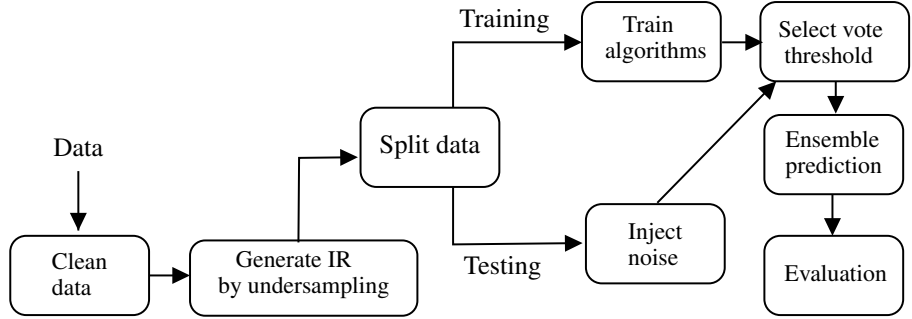

The experimental protocol adopted in this work is summarized in Figure~\ref{fig:graph-protocol}. 
The protocol starts with a real-world dataset given as input. Initially, a data cleaning process using the consensus vote is applied to remove possible noise from the input dataset. Then a new dataset is generated from the cleaned data with the desired class imbalance ratio ($IR$). The generated dataset is split into training (70\%) and testing (30\%) data. The training data is used to produce the pool of classifiers used for ensemble noise detection. 

For evaluating the noise detector, the percentage of noise \emph{p} is injected into the testing data according to the noise ratio \emph{M} and a chosen noise model. Each test instance is given as input to the pool of classifiers. With all predictions, the test instance is marked as noisy if at least \emph{L} classifiers misclassify it. Then, some evaluation measures described in this section are calculated and analyzed.

The main purpose of the proposed methodology of experiments is to derive insights on how to design ensemble noise detectors under specific conditions, like different class distributions and noise levels per class. It is important to highlight that domain knowledge and experts (when available) can be very helpful to estimate such specific conditions. For instance the noise level per class can be estimated by relying on an expert inspection of a sample of data instances to identify eventual labeling errors. Once such conditions are identified via auxiliary data and experts, the design of the noise detectors can be more adequately performed.

\FloatBarrier

In Section~\ref{sec:setup_ensemble}, the algorithms for the ensemble filter and the vote scheme approach are presented. In Section~\ref{sec:setup_datasets}, the real-world datasets are detailed, and it is described the methodology for data generation with specific settings. The procedure for noise injection regarding the noise model is explained in Section~\ref{sec:setup_noise_injection}. Depending on the label noise model adopted in the experiments, noise detection techniques can have distinct performance behavior, usually measured by precision and recall, defined in Section~\ref{sec:background_performance_measure}. Lastly, the input variables and the experimental protocol are summarized in Section~\ref{sec:setup_protocol}.

\subsection{Ensemble}
\label{sec:setup_ensemble}

The noise detection ensemble used in the experiments was generated from 10 algorithms adopted in the related work  \citep{Sluban-diversity}. They were chosen from different families: decision trees, Bayesian models, neural networks, support vector machines, random forest, nearest neighbors, and ruled-based methods, forming a diverse pool of classifiers.  All of them are implemented in R from specific packages, as shown in Table \ref{tab:algorithms}. Default parameters suggested in the R packages were employed for all algorithms in the experiments. Parameters could be optimized for each algorithm, which would result in more accurate classifiers. However, as we are dealing with ensembles, the lack of parameter optimization would be compensated by using diverse algorithms. The impact of parameter optimization in ensemble noise detection will be better investigated in future work.

\begin{table}[h]
 \setlength\extrarowheight{2pt}
\caption{Learning algorithms for classification noise filtering.}
\label{tab:algorithms}
\begin{center}
\begin{tabularx}{0.7\textwidth}{ll}
\toprule
\textbf{Algorithm} & \textbf{R Package}  \\ \hline
CN2 (rule learner)                     & RoughSets                               \\
kNN (nearest neighbor)                 & class                                   \\
Naive Bayes                            & naivebayes                              \\
Random forest                          & randomForest                            \\
SVM (RBF Kernel)                       & e1071                                   \\
J48                                    & RWeka                                   \\
JRip                                   & RWeka                                   \\
Multiplayer perceptron                 & RSNNS                                   \\
Decision tree                          & party                                   \\
SMO (linear Kernel)                    & RWeka                                   \\
\bottomrule
\end{tabularx}

\end{center}
\end{table}

\pagebreak

Unlike previous work that adopted the majority and the consensus vote approaches for ensemble-based noise filtering, in our work, we evaluate different values for the decision threshold \emph{L} in the ensemble. If more than \emph{L}\% of classifiers in the pool misclassify an instance, it is considered wrongly labeled. In our experiments, the threshold \emph{L} varies from 10\% to 100\%. The majority and the consensus approaches are special cases, respectively, assuming \emph{L} = 50\% (i.e., more than half classifiers plus 1) and 100\% (i.e., all classifiers). Of course, too small values, like 10\%, would be reasonable only in contexts with almost noise-free datasets and/or ones that want to force a very high level of recall in noise detection. Nevertheless, we vary \emph{L} to check if there is a threshold that maximizes the noise detection under specific contexts, i.e., considering the class imbalance ratio, noise distribution, and percentage of total noise.
 
\subsection{Data}
\label{sec:setup_datasets}

Quantitative assessment of noise detection methods requires knowing which are the noisy instances beforehand. In real-world datasets, this is achieved either by expert labeling or by randomly injecting artificial noise into a dataset. While the former approach is not feasible for an extensive evaluation, the latter still has uncertainty about which instances are originally noisy when dealing with real-world datasets. 
\addtocounter{table}{-1} 

\begin{table}[h]
 \setlength\extrarowheight{2pt}
\caption{Real-world data information. \textit{Missing} = instances with at least one missing value.}
\label{tab:dataset_info}
\begin{center}

\begin{tabularx}{0.7\textwidth}{l|r|ccc|ccc}
\toprule
\multirow{2}{*}{\textbf{Dataset}}  & \multirow{2}{*}{\textbf{Attr.}}  & \multicolumn{3}{c|}{\textbf{Original}} & \multicolumn{3}{c}{\textbf{After cleaning}} \\ \cline{3-8} 
								   &                            &            \textbf{IR}  & \textbf{Inst.} & \textbf{Missing} & \textbf{IR}  & \textbf{Inst.} & \textbf{ Removal (\%)} \\ \hline
								   
arcene & 10001 & 44:56 & 200 & 0 & 44:56 & 199 & 0.50\\ 
breast-c & 10 & 34:66 & 699 & 16 & 35:65 & 673 & 3.72\\ 
column2C & 7 & 32:68 & 310 & 0 & 32:68 & 308 & 0.65\\ 
credit & 16 & 44:56 & 690 & 37 & 45:55 & 644 & 6.67\\ 
cylinder-bands & 40 & 42:58 & 540 & 263 & 36:64 & 276 & 48.89\\ 
diabetes & 9 & 35:65 & 768 & 0 & 31:69 & 720 & 6.25\\ 
eeg-eye-state & 15 & 45:55 & 14980 & 0 & 45:55 & 14979 & 0.01\\ 
glass0 & 10 & 33:67 & 214 & 0 & 33:67 & 214 & 0.00\\ 
glass1 & 10 & 36:64 & 214 & 0 & 35:65 & 212 & 0.93\\ 
heart-c & 14 & 46:54 & 303 & 7 & 45:55 & 289 & 4.62\\ 
heart-statlog & 14 & 44:56 & 270 & 0 & 44:56 & 262 & 2.96\\ 
hill-valley & 101 & 50:50 & 1212 & 0 & 48:52 & 1184 & 2.31\\ 
ionosphere & 35 & 36:64 & 351 & 0 & 35:65 & 345 & 1.71\\ 
kr-vs-kp & 37 & 48:52 & 3196 & 0 & 48:52 & 3194 & 0.06\\ 
mushroom & 23 & 48:52 & 8124 & 2480 & 38:62 & 5644 & 30.53\\ 
pima & 9 & 35:65 & 768 & 0 & 32:68 & 732 & 4.69\\ 
sonar & 61 & 47:53 & 208 & 0 & 46:54 & 206 & 0.96\\ 
steel-plates-fault & 34 & 35:65 & 1941 & 0 & 35:65 & 1941 & 0.00\\ 
tic-tac-toe & 10 & 35:65 & 958 & 0 & 35:65 & 958 & 0.00\\ 
voting & 17 & 39:61 & 435 & 203 & 47:53 & 228 & 47.59\\

\bottomrule					
\end{tabularx}

\end{center}
\end{table}


In our experiments, we adopted 20 real-world binary datasets available at the KEEL-dataset repository \citep{KEEL}, UCI repository  \citep{UCI:2017}, and Open Media Library \citep{OpenML2013}. Some multi-class datasets are modified to obtain two-class imbalanced problems, defining the joint of one or more classes as positive and the remainder as negative. The list of datasets is presented in Table \ref{tab:dataset_info}.

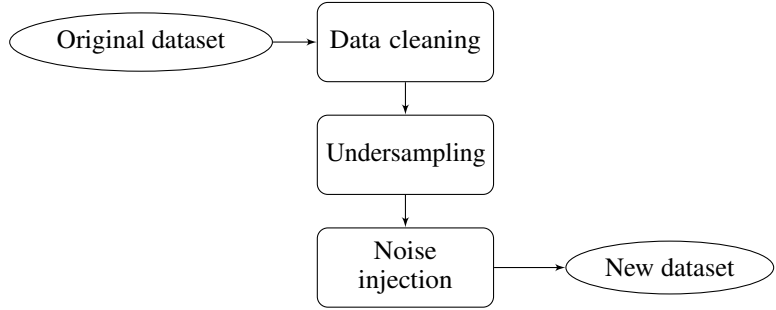
\begin{figure}[ht]
\begin{center}
\tikzstyle{block} = [rectangle, draw,  
    text width=6em, text centered, rounded corners, minimum height=3em]
\tikzstyle{line} = [draw, -latex']
\tikzstyle{cloud} = [draw, ellipse,fill=red!20, node distance=3cm, minimum height=2em]
\tikzstyle{square_block} = [ellipse, draw, text centered,node distance=10em,  minimum height=2em]
    
\begin{tikzpicture}[node distance = 1.5cm, auto]
    \node [block] (init) {Data cleaning};
    \node [square_block,left of=init] (original) {Original dataset};
    \node [block, below of=init] (identify) {Undersampling};
    \node [block, below of=identify] (evaluate) {Noise injection};
    \node [square_block,right of=evaluate] (newData) {New dataset};
    \path [line] (original) -- (init);
    \path [line] (init) -- (identify);
    \path [line] (identify) -- (evaluate);
     \path [line] (evaluate) -- (newData);
\end{tikzpicture}
\caption{Data generation process.}
\label{fig:dataset_gen_process}
\end{center}
\end{figure}
\FloatBarrier

For each dataset, a three-step process is adopted (see Figure \ref{fig:dataset_gen_process}) to generate a new dataset with controlled IR and injected label noise. The dataset generation process is described below:

\begin{itemize}

    \item \textbf{Data cleaning:} 
    As suggested in \cite{Sluban-diversity}, a data cleaning step is applied to reduce the inherent label noise that is possibly present in the dataset. Hence, the evaluation of noise detection will be mainly contingent on the artificial noise injected in a controlled manner. In this step, a 10-fold classification is employed, and the consensus method is used to remove noisy instances. The consensus vote was chosen for this step for being more strict, ensuring that only instances undoubtedly noisy are discarded since it requires all ensemble classifiers' agreement. 

    \item  \textbf{Undersampling:}
    In this step, three datasets are generated for each cleaned data, in order to obtain the following {IR} configurations: (1) 50:50, (2) 30:70, and (3) 20:80. For generating a dataset with a given IR, a random undersampling process of the majority class is applied.
    
    \item \textbf{Noise injection:} In this step, noise is artificially injected according to a given label noise model. This step is detailed in Section \ref{sec:setup_noise_injection}.
\end{itemize}

\subsection{Noise Injection}
\label{sec:setup_noise_injection}

For noise detection evaluation, label noise is injected into the test set by changing the class label in a determined proportion $p$ of records. In our experiments, the desired noise level \emph{p} assumed four different values: $5\%, 10\%, 15\%$, and $20\%$ of instances. 

For the NAR model, the noise is inserted to achieve a specific ratio $M$ of noisy instances per class. Two values of $M$ were chosen in such a way to analyze the impact of a significant discrepancy in the noise level per class: 
\vspace{0.1in}

\begin{itemize}

\item $M = 9/1$: for every nine noisy instances in the minority class, there is one noisy instance in the majority class. It simulates a scenario of application in which it is more difficult to label examples in the minority class. This chosen ratio is referred to as NAR (9:1) in the discussion of results. 

\item $M = 1/9$: in turn, for each noisy instance in the minority class, there are nine noisy instances in the majority class. It means that the majority class is more prone to have label noise. This chosen ratio corresponds to NAR (1:9) in the discussion of results.  

\end{itemize}

 Notice that the NCAR model is a particular case of NAR by assuming $M = 1$, i.e., NCAR is equivalent to NAR (1:1).

Each class's exact number of noisy instances is determined according to the desired noise level $p$ and the ratio $M$. Let $d_n$ be the number of instances in the test set. Let $n_1$ and $n_2$ be the number of noisy instances in each class. Then $n_1 + n_2 = p\times d_n$ and $M = n_1/n_2$. In order to obtain such constraints, $n_1$ and $n_2$ are defined according to the following equations:
 
\begin{equation}
 n_1 = \frac{M\times (d_n\times p)}{M+1}
\end{equation}

\begin{equation}
n_2 = \frac{d_n\times p}{M+1}
\end{equation}

\begin{figure}[ht]
  \centering
  \includegraphics[width=5.2in]{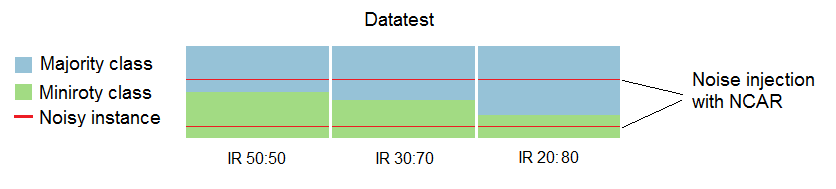}
  \caption{Noise injection with NCAR model for different imbalance ratios (IR).}
  \label{fig:noise_injection_NCAR}
\end{figure}

For example, suppose that we have $d_n = 1000$ records in the test set, and the desired noise level is $p=0.10$ (100 noisy instances). Then, consider that noise is injected according to the three settings: NCAR, NAR (9:1) NAR (1:9). In the first setting, $M=1$, and by adopting the above equations, we would find $n_1 = 50 $ and $n_2 = 50$. The number of noisy instances in the test set is the same for both classes. This example is illustrated in Figure~\ref{fig:noise_injection_NCAR}.

 \FloatBarrier
\FloatBarrier

\begin{figure}[ht]
  \centering
  \includegraphics[width=5.2in]{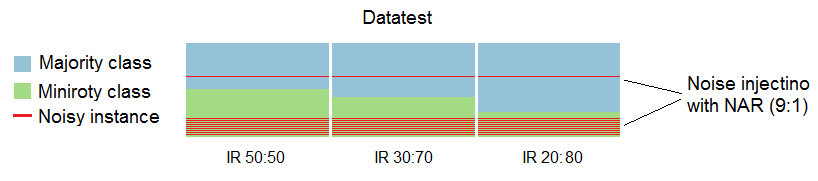}
  \caption{Noise injection with NAR 9:1 model for different imbalance ratios (IR).}
  \label{fig:noise_injection_NAR91}
\end{figure}

In the NAR (9:1) setting, $M=9$ and then $n_1 = 90$ (90 noisy instances in minority class) and $n_2 = 10$ (10 noisy instances in majority class). In this case, the noise level in minority class is higher in the test set. This example is illustrated in Figure~\ref{fig:noise_injection_NAR91}.

 \FloatBarrier
\FloatBarrier

\begin{figure}[ht]
  \centering
  \includegraphics[width=5.2in]{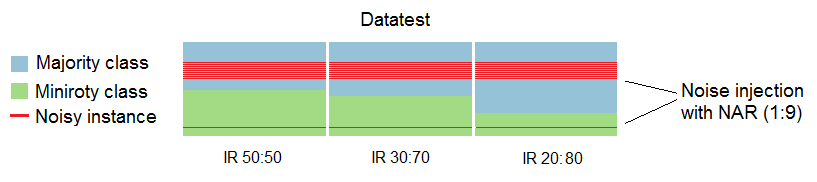}
  \caption{Noise injection with NAR 1:9 model for different imbalance ratios (IR).}
  \label{fig:noise_injection_NAR19}
\end{figure}

Finally, for the NAR (1:9) setting, $M=1/9$, hence, $n_1 = 10$ (10 noisy instances in the minority class) and $n_2 = 90$ (90 noisy instances in the majority class). In this case, the number of noisy instances injected in the test set is greater for the majority class, as illustrated in Figure~\ref{fig:noise_injection_NAR19}.

\FloatBarrier
\FloatBarrier

\subsection{Performance Measures} \label{sec:background_performance_measure}
Most experiments in the literature assess the efficiency of methods in detecting noise regarding accuracy \citep{survey}. A primary measure to evaluate the performance of noise detection is \textit{precision}, which means how many noisy instances the detector correctly identified among all records identified as noisy:

\[
  \mbox{Precision}  = \frac{\mbox{number of noisy cases correctly identified}}{\mbox{number of all noisy cases identified}}
\]

In addition to the precision, another helpful measure is \textit{recall}, which calculates how many instances the detector correctly identified as noisy among all the noisy records inserted into the dataset:

\[ 
  \mbox{Recall}  = \frac{\mbox{number of noisy cases correctly identified}}{\mbox{number of all noisy cases in dataset}}
\]

Finally, a measure that trades off \textit{precision} versus \textit{recall} is the \textit{F-score}, which is the weighted harmonic mean of \textit{precision} and \textit{recall}: 

\begin{equation}
\textit{F-score} = \frac{\beta\times Precision \times Recall}{Precision + Recall} \label{eq_precision}
\end{equation}

\noindent where $\beta^2 = \frac{1-\alpha}{\alpha}$, with $\alpha \in [0,1]$ and $\beta^2 \in [0, \infty]$.

Setting the $\beta$  parameter makes it possible to assign more importance to either precision or recall when calculating the \textit{F-score}. In this work, the standard \textit{F-score} (also referred to as $F_{1} score$, $\beta = 1$ and $\alpha = 1/2$ ) was used. It equally weights precision and recall. 


\subsection{Experimental Protocol Algorithm}
\label{sec:setup_protocol}

The experimental protocol illustrated in Figure~\ref{fig:graph-protocol} is also outlined in Algorithm~\ref{algo:exp_procedure_real}.  The algorithm was executed for each input parameter combination shown in Table~\ref{tab:exp-strucuture}. Given the stochastic nature of noise injection, this insertion is usually repeated several times for each noise level \citep{Sluban2014} \citep{zhu-large}. In this work, for each  parameters combination, Algorithm~\ref{algo:exp_procedure_real} was repeated 100 times, and the average results of \textit{Precision}, \textit{Recall}, and \textit{F1-Measure} were employed to evaluate the ensemble noise detector.

\addtocounter{table}{-1} 
\begin{table}[t]
 \setlength\extrarowheight{2pt}
\caption{Experimental setup}
\label{tab:exp-strucuture}
\begin{center}
\begin{tabularx}{0.7\textwidth}{c|c|c|c|c|l}
\toprule

Imbalance Ratio (IR)      &  \multicolumn{4}{c|}{Total percentage of}               & Noise Ratio (M)    \\ \cline{1-1} \cline{6-6}
Class min : Class maj         &   \multicolumn{4}{c|}{noise in testing set (p)}                                                                       & Class min : Class maj  \\ \hline
\multirow{3}{*}{50 : 50 } & \multirow{3}{*}{5\%} & \multirow{3}{*}{10\%} & \multirow{3}{*}{15\%} & \multirow{3}{*}{20\%} & NCAR               \\  
                          &                      &                       &                       &                       & NAR (1 : 9)        \\  
                          &                      &                       &                       &                       & NAR (9 : 1)        \\ \hline
\multirow{3}{*}{30 : 70}  & \multirow{3}{*}{5\%} & \multirow{3}{*}{10\%} & \multirow{3}{*}{15\%} & \multirow{3}{*}{20\%} & NCAR               \\  
                          &                      &                       &                       &                       & NAR (1 : 9)        \\  
                          &                      &                       &                       &                       & NAR (9 : 1)        \\ \hline
\multirow{3}{*}{20 : 80}  & \multirow{3}{*}{5\%} & \multirow{3}{*}{10\%} & \multirow{3}{*}{15\%} & \multirow{3}{*}{20\%} & NCAR               \\  
                          &                      &                       &                       &                       & NAR (1 : 9)        \\  
                          &                      &                       &                       &                       & NAR (9 : 1)        \\ 
\bottomrule
\end{tabularx}

\end{center}
\end{table}

\begin{algorithm}
\caption{Experimental procedure}
\label{algo:exp_procedure_real}
\begin{center}
\begin{flushleft}
\textbf{Input:} IR, M, p, dataset, classifiers, threshold (L)
\\
\textbf{Output: } performance measures
\end{flushleft}
\begin{algorithmic}[1]
\STATE $cleanData\gets dataCleasing(dataset,classifiers)$
\STATE $data\gets generateData(cleanData, IR)$
\STATE $training, testing\gets split(data, 70\%,30\%)$ \COMMENT{Split data in training and testing data}
\STATE $noisy\_testing\gets injectNoise(testing,M,p)$
\FOR{c in classifiers} 
\STATE $model\gets train(training,c)$
\STATE $predictions\gets classify(model,noisy\_testing)$
\ENDFOR
\STATE $ensemble\_prediction\gets voting(predictions,L)$
\STATE $measures\gets calculate(ensemble\_prediction)$
\end{algorithmic}
\end{center}
\end{algorithm}


\section{Results}
\label{sec:results}

In this section, the findings from the experiments described in Section~\ref{sec:setup_protocol} are presented and examined. The results shown in this section were obtained from the steps outlined in Algorithm~\ref{algo:exp_procedure_real}. The discussion is carried out by putting into perspective each input parameter that resulted in a specific scenario, facilitating the analysis and comparison. In this way, we first examine the noise detection concerning balanced and imbalanced datasets exploring the impact of different noise levels per class in Section~\ref{sec:real_balancec_vs_imbalanced}. In Section~\ref{sec:real_different_thresholds}, we analyze noise detection under different noise ensemble thresholds. Lastly, statistical tests of the results are detailed in Section~\ref{sec:stat_tests}.

\subsection{Imbalance Ratio and Noise Level per Class}
\label{sec:real_balancec_vs_imbalanced}

\begin{figure}[h!]
  \centering
  \includegraphics[width=5.5in]{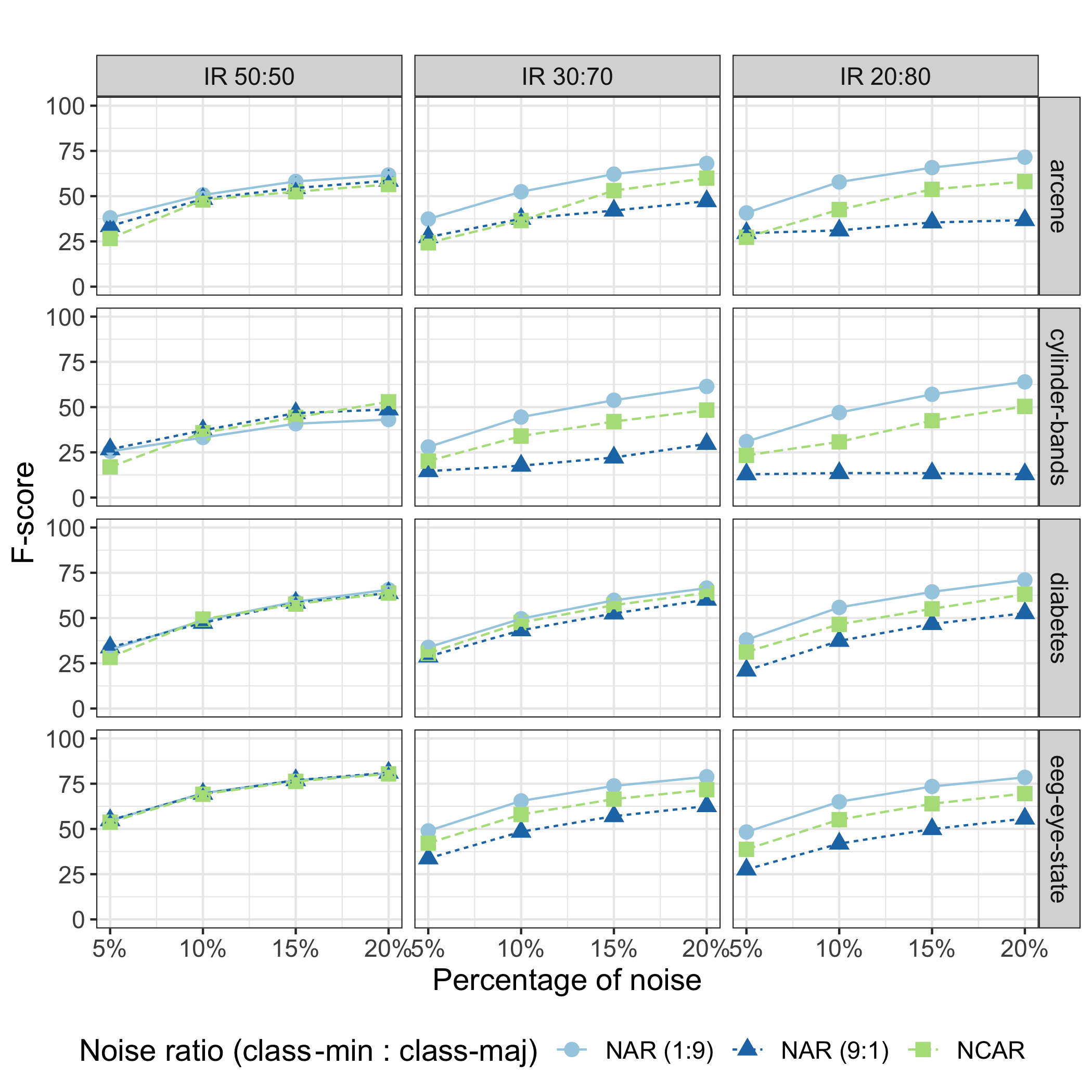}
 \caption{\textit{F-score} performance for majority vote on \emph{arcene}, \emph{cylinder-bands}, \emph{diabetes}, and \emph{eeg-eye-state} datasets.}
 \label{fig:results_real_fscore_comparison_1}
\end{figure}

\begin{figure}[h!]
  \centering
  \includegraphics[width=5.5in]{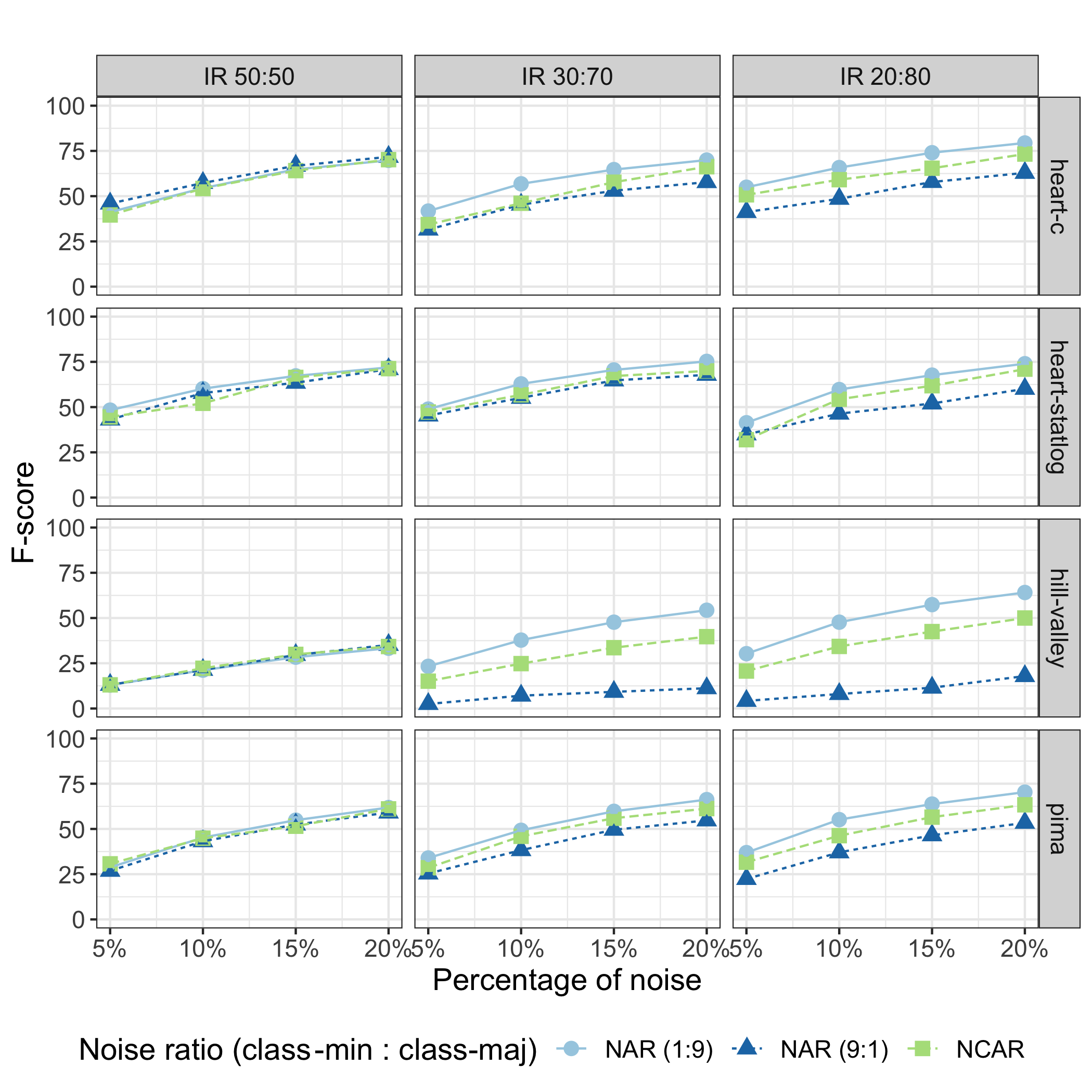}
 \caption{\textit{F-score} performance for majority vote on \emph{heart-c}, \emph{heart-statlog}, \emph{hill-valley}, and \emph{pima} datasets.}
 \label{fig:results_real_fscore_comparison_2}
\end{figure}

Figures~\ref{fig:results_real_fscore_comparison_1} and \ref{fig:results_real_fscore_comparison_2} show the {\it F-score} for eight datasets, varying the noise level and the imbalance ratio. F-score tends to increase as the general noise level (p) increases as well. Similar patterns of results were observed in the other datasets. This is expected since the noise detection task becomes easier for higher amounts of noise in a dataset.

\FloatBarrier
\FloatBarrier

\FloatBarrier
\FloatBarrier

The general behavior observed in 
the  scenario of balanced data (IR 50:50 - first column) was that NCAR and NAR models produced the same effect on noise detection regardless of the noise distribution (noise level per class). In other terms, when the dataset is balanced, the noise detection performance depends more on the percentage of noise in the dataset than on how the noise is distributed per class. 

For imbalanced datasets, noise detection is affected by the noise level per class, as revealed by the F-score discrepancy between NCAR and NAR models shown in Figures~\ref{fig:results_real_fscore_comparison_1} and \ref{fig:results_real_fscore_comparison_2}. This difference increases when the noise level gets higher, as observed for \textit{arcene} and \textit{cylinder-bands} datasets in a more pronounced way. Smaller differences in NCAR and NAR models results were mainly observed when the noise level was 5\%. In such a noise level, it is, in fact, more difficult to obtain good performance measures.

\addtocounter{table}{-1} 
\begin{table}[h!] 
 \setlength\extrarowheight{2pt}
\caption{F-score variation vs class imbalance ratio.}
\label{tab:fscoreVariationRealData}
\begin{center}

\resizebox{\columnwidth}{!}{ %
\begin{tabular}{l|rrr|rrr|rrr|rrr}
\toprule
\multirow{3}{*}{\textbf{Datasets}}  &  \multicolumn{12}{c}{\textbf{F-score variation when IR goes from 50:50 to 20:80}}                                                                                                  \\ \cline{2-13} 
                                    &  \multicolumn{3}{c|}{\textbf{5\% of noise}}                                                                                                   & \multicolumn{3}{c|}{\textbf{10\% of noise}}                                                                                                   & \multicolumn{3}{c|}{\textbf{15\% of noise}}                                                                                                    & \multicolumn{3}{c}{\textbf{20\% of noise}}                                                                                                   \\ \cline{2-13} 
                                    &  \textbf{\begin{tabular}[c]{@{}c@{}}NAR\\ (9:1)\end{tabular}} & \textbf{NCAR} & \textbf{\begin{tabular}[c]{@{}c@{}}NAR\\ (1:9)\end{tabular}} & \textbf{\begin{tabular}[c]{@{}c@{}}NAR\\ (9:1)\end{tabular}} & \textbf{NCAR} & \textbf{\begin{tabular}[c]{@{}c@{}}NAR\\ (1:9)\end{tabular}} & \textbf{\begin{tabular}[c]{@{}c@{}}NAR\\ (9:1)\end{tabular}} & \textbf{NCAR} & \textbf{\begin{tabular}[c]{@{}c@{}}NAR\\ (1:9)\end{tabular}} & \textbf{\begin{tabular}[c]{@{}c@{}}NAR\\ (9:1)\end{tabular}} & \textbf{NCAR} & \textbf{\begin{tabular}[c]{@{}c@{}}NAR\\ (1:9)\end{tabular}} \\ \hline

arcene & \textcolor{red}{-3.6} & \textcolor{blue}{1.1} & \textcolor{blue}{2.8} & \textcolor{red}{-17.4} & \textcolor{red}{-5.4} & \textcolor{blue}{7.1} & \textcolor{red}{-19.2} & \textcolor{blue}{1.3} & \textcolor{blue}{7.6} & \textcolor{red}{-21.6} & \textcolor{blue}{1.7} & \textcolor{blue}{9.8} \\ 
breast-cancer-wisconsin & \textcolor{red}{-1.1} & \textcolor{red}{-1.4} & \textcolor{blue}{4.7} & \textcolor{red}{-0.4} & \textcolor{black}{0.0} & \textcolor{blue}{3.8} & \textcolor{red}{-0.9} & \textcolor{red}{-0.6} & \textcolor{blue}{2.2} & \textcolor{red}{-0.4} & \textcolor{red}{-0.1} & \textcolor{blue}{1.7} \\ 
column2C & \textcolor{red}{-13.6} & \textcolor{blue}{4.1} & \textcolor{red}{-2.8} & \textcolor{red}{-10.5} & \textcolor{red}{-2.1} & \textcolor{red}{-0.4} & \textcolor{red}{-15.6} & \textcolor{blue}{1.7} & \textcolor{blue}{1.8} & \textcolor{red}{-13.3} & \textcolor{red}{-2.4} & \textcolor{blue}{0.9} \\ 
credit & \textcolor{red}{-5.8} & \textcolor{red}{-1.2} & \textcolor{blue}{2.3} & \textcolor{red}{-8.6} & \textcolor{blue}{0.6} & \textcolor{blue}{1.6} & \textcolor{red}{-7.0} & \textcolor{black}{0.0} & \textcolor{blue}{1.4} & \textcolor{red}{-6.6} & \textcolor{red}{-0.3} & \textcolor{blue}{2.0} \\ 
cylinder-bands & \textcolor{red}{-13.8} & \textcolor{blue}{6.6} & \textcolor{blue}{5.4} & \textcolor{red}{-23.6} & \textcolor{red}{-5.1} & \textcolor{blue}{13.8} & \textcolor{red}{-33.0} & \textcolor{red}{-1.8} & \textcolor{blue}{16.2} & \textcolor{red}{-35.8} & \textcolor{red}{-2.6} & \textcolor{blue}{20.8} \\ 
diabetes & \textcolor{red}{-12.8} & \textcolor{blue}{3.0} & \textcolor{blue}{5.5} & \textcolor{red}{-10.0} & \textcolor{red}{-2.9} & \textcolor{blue}{7.2} & \textcolor{red}{-11.8} & \textcolor{red}{-2.6} & \textcolor{blue}{5.4} & \textcolor{red}{-11.1} & \textcolor{red}{-0.6} & \textcolor{blue}{5.4} \\ 
eeg-eye-state & \textcolor{red}{-27.2} & \textcolor{red}{-15.0} & \textcolor{red}{-6.3} & \textcolor{red}{-27.5} & \textcolor{red}{-14.0} & \textcolor{red}{-4.8} & \textcolor{red}{-27.0} & \textcolor{red}{-12.4} & \textcolor{red}{-3.4} & \textcolor{red}{-25.4} & \textcolor{red}{-10.9} & \textcolor{red}{-2.2} \\ 
glass0 & \textcolor{red}{-1.5} & \textcolor{blue}{1.9} & \textcolor{red}{-3.0} & \textcolor{red}{-0.5} & \textcolor{blue}{8.8} & \textcolor{blue}{7.9} & \textcolor{red}{-5.1} & \textcolor{blue}{3.5} & \textcolor{blue}{6.5} & \textcolor{red}{-4.5} & \textcolor{blue}{3.3} & \textcolor{blue}{8.9} \\ 
glass1 & \textcolor{red}{-7.6} & \textcolor{red}{-6.2} & \textcolor{red}{-0.8} & \textcolor{red}{-18.1} & \textcolor{blue}{3.1} & \textcolor{blue}{8.8} & \textcolor{red}{-21.6} & \textcolor{blue}{2.2} & \textcolor{blue}{8.9} & \textcolor{red}{-27.0} & \textcolor{red}{-1.8} & \textcolor{blue}{11.9} \\ 
heart-c & \textcolor{red}{-4.7} & \textcolor{blue}{11.1} & \textcolor{blue}{13.7} & \textcolor{red}{-8.8} & \textcolor{blue}{4.9} & \textcolor{blue}{11.4} & \textcolor{red}{-9.0} & \textcolor{blue}{1.3} & \textcolor{blue}{9.2} & \textcolor{red}{-8.8} & \textcolor{blue}{3.1} & \textcolor{blue}{9.6} \\ 
heart-statlog & \textcolor{red}{-8.2} & \textcolor{red}{-12.6} & \textcolor{red}{-7.0} & \textcolor{red}{-11.4} & \textcolor{blue}{2.3} & \textcolor{red}{-0.5} & \textcolor{red}{-11.4} & \textcolor{red}{-4.6} & \textcolor{blue}{0.4} & \textcolor{red}{-10.9} & \textcolor{red}{-0.3} & \textcolor{blue}{2.1} \\ 
hill-valley & \textcolor{red}{-8.8} & \textcolor{blue}{7.6} & \textcolor{blue}{17.4} & \textcolor{red}{-13.3} & \textcolor{blue}{12.1} & \textcolor{blue}{26.5} & \textcolor{red}{-18.1} & \textcolor{blue}{12.6} & \textcolor{blue}{29.0} & \textcolor{red}{-17.2} & \textcolor{blue}{15.7} & \textcolor{blue}{30.8} \\ 
ionosphere & \textcolor{red}{-4.6} & \textcolor{red}{-5.3} & \textcolor{blue}{3.6} & \textcolor{red}{-4.1} & \textcolor{red}{-2.8} & \textcolor{blue}{3.3} & \textcolor{red}{-6.4} & \textcolor{blue}{1.5} & \textcolor{blue}{1.1} & \textcolor{red}{-4.7} & \textcolor{blue}{0.1} & \textcolor{blue}{1.9} \\ 
kr-vs-kp & \textcolor{red}{-8.7} & \textcolor{red}{-6.8} & \textcolor{red}{-7.4} & \textcolor{red}{-5.9} & \textcolor{red}{-3.9} & \textcolor{red}{-3.6} & \textcolor{red}{-4.7} & \textcolor{red}{-3.0} & \textcolor{red}{-3.0} & \textcolor{red}{-3.9} & \textcolor{red}{-2.1} & \textcolor{red}{-2.0} \\ 
mushroom & \textcolor{red}{-0.1} & \textcolor{red}{-0.1} & \textcolor{red}{-0.1} & \textcolor{black}{-0.0} & \textcolor{black}{-0.0} & \textcolor{red}{-0.1} & \textcolor{black}{-0.0} & \textcolor{black}{-0.0} & \textcolor{black}{-0.0} & \textcolor{black}{-0.0} & \textcolor{black}{-0.0} & \textcolor{black}{-0.0} \\ 
pima & \textcolor{red}{-4.5} & \textcolor{blue}{0.9} & \textcolor{blue}{8.2} & \textcolor{red}{-6.1} & \textcolor{blue}{1.4} & \textcolor{blue}{9.9} & \textcolor{red}{-5.9} & \textcolor{blue}{5.0} & \textcolor{blue}{8.9} & \textcolor{red}{-5.7} & \textcolor{blue}{2.3} & \textcolor{blue}{8.5} \\ 
sonar & \textcolor{blue}{14.1} & \textcolor{blue}{18.6} & \textcolor{blue}{15.0} & \textcolor{blue}{13.4} & \textcolor{blue}{13.4} & \textcolor{blue}{19.4} & \textcolor{blue}{9.0} & \textcolor{blue}{13.7} & \textcolor{blue}{17.8} & \textcolor{blue}{6.3} & \textcolor{blue}{10.0} & \textcolor{blue}{16.5} \\ 
steel-plates-fault & \textcolor{black}{0.0} & \textcolor{blue}{1.0} & \textcolor{blue}{0.4} & \textcolor{black}{-0.0} & \textcolor{blue}{0.4} & \textcolor{blue}{0.3} & \textcolor{black}{0.0} & \textcolor{blue}{0.4} & \textcolor{blue}{0.2} & \textcolor{black}{-0.0} & \textcolor{blue}{0.1} & \textcolor{blue}{0.1} \\ 
tic-tac-toe & \textcolor{red}{-22.1} & \textcolor{red}{-4.9} & \textcolor{blue}{5.0} & \textcolor{red}{-26.1} & \textcolor{red}{-5.9} & \textcolor{blue}{5.4} & \textcolor{red}{-26.6} & \textcolor{red}{-5.2} & \textcolor{blue}{5.1} & \textcolor{red}{-27.1} & \textcolor{red}{-5.4} & \textcolor{blue}{5.4} \\ 
voting & \textcolor{red}{-6.8} & \textcolor{red}{-13.6} & \textcolor{red}{-4.1} & \textcolor{red}{-5.8} & \textcolor{red}{-7.6} & \textcolor{red}{-2.5} & \textcolor{red}{-4.4} & \textcolor{red}{-6.2} & \textcolor{red}{-2.0} & \textcolor{red}{-4.7} & \textcolor{red}{-4.8} & \textcolor{red}{-1.4} \\

\bottomrule
\end{tabular}
}
\end{center}
\end{table}

Table~\ref{tab:fscoreVariationRealData} shows how F-score changes when the scenario varies from a balanced dataset to an imbalanced dataset. Negative and positive numbers denote, respectively, a decrease and increase in noise detection. As can be seen, noise distribution is important when considered in combination with class imbalance. For instance, under the NAR (9:1) model (i.e., more noise instances in the minority class than in the majority's), noise detection worsened its performance when class imbalance was increased. This was the overall behavior confirmed by the negative F-score variation presented in Table~\ref{tab:fscoreVariationRealData} at NAR (9:1) column. On the other hand, an opposite pattern of results was observed for the NAR (1:9) model, i.e., noise detection was improved when class imbalance increased. The greater number of positive F-score variations found at the NAR (1:9) column endorse that, in imbalanced datasets, noisy instances in the majority class tend to be more easily detected than noisy instances in the minority class. No consistent pattern of performance was observed in Table~\ref{tab:fscoreVariationRealData} for the NCAR model. This indicates that when the noise is evenly distributed in an imbalanced dataset, the particularities and difficulties of the problem itself may be more crucial in noise detection performance than the IR.

\begin{figure}[!]
  \centering
  \includegraphics[width=5.5in]{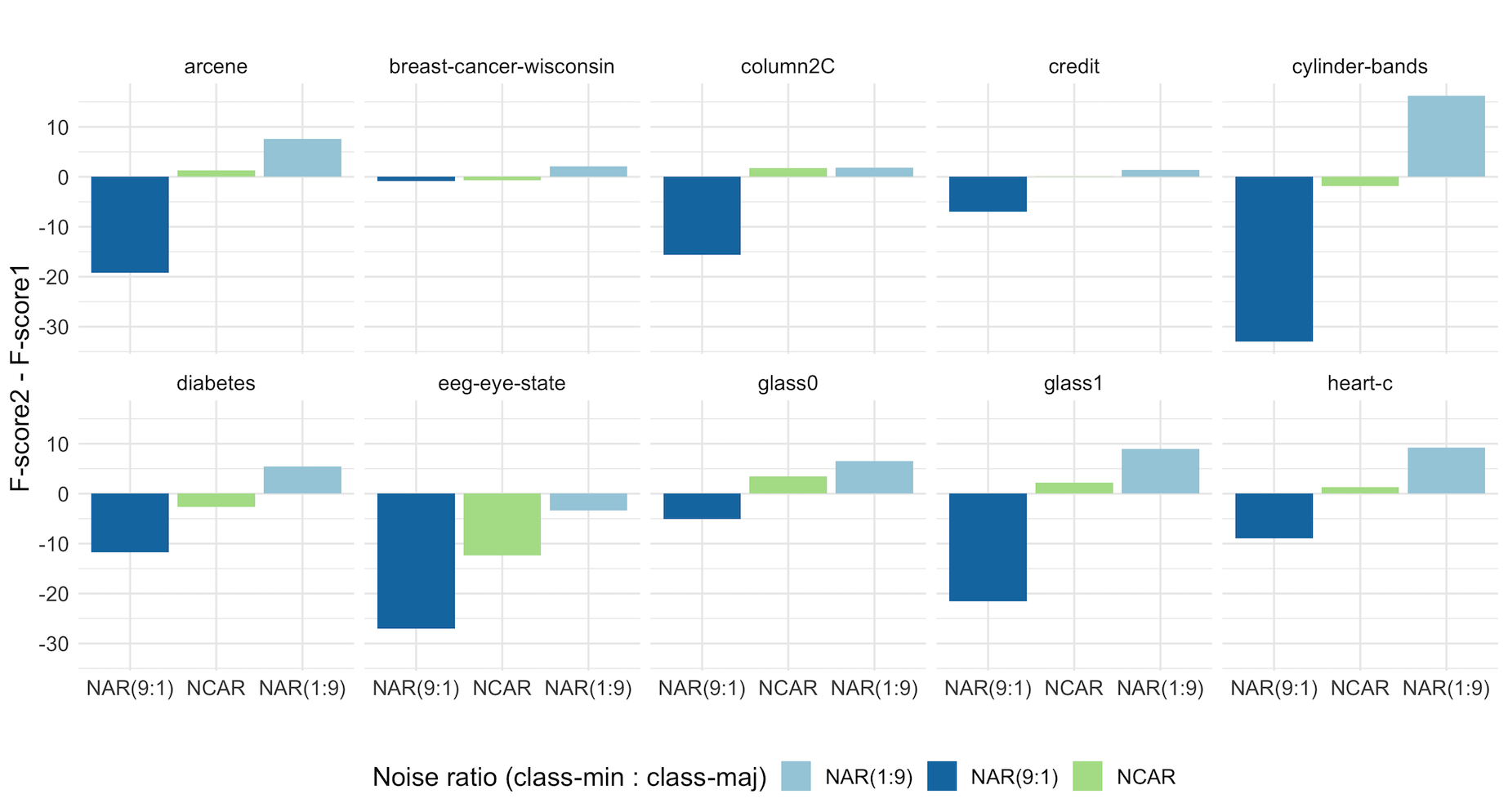}
 \caption{Variation in \textit{F-score} performance when IR increases from 50:50 (F-score1) to 20:80 (F-score2) in presence of a noise level of 15\%.}
 \label{fig:results_real_fscore_variation_15_1}
\end{figure}

\begin{figure}
  \centering
  \includegraphics[width=5.5in]{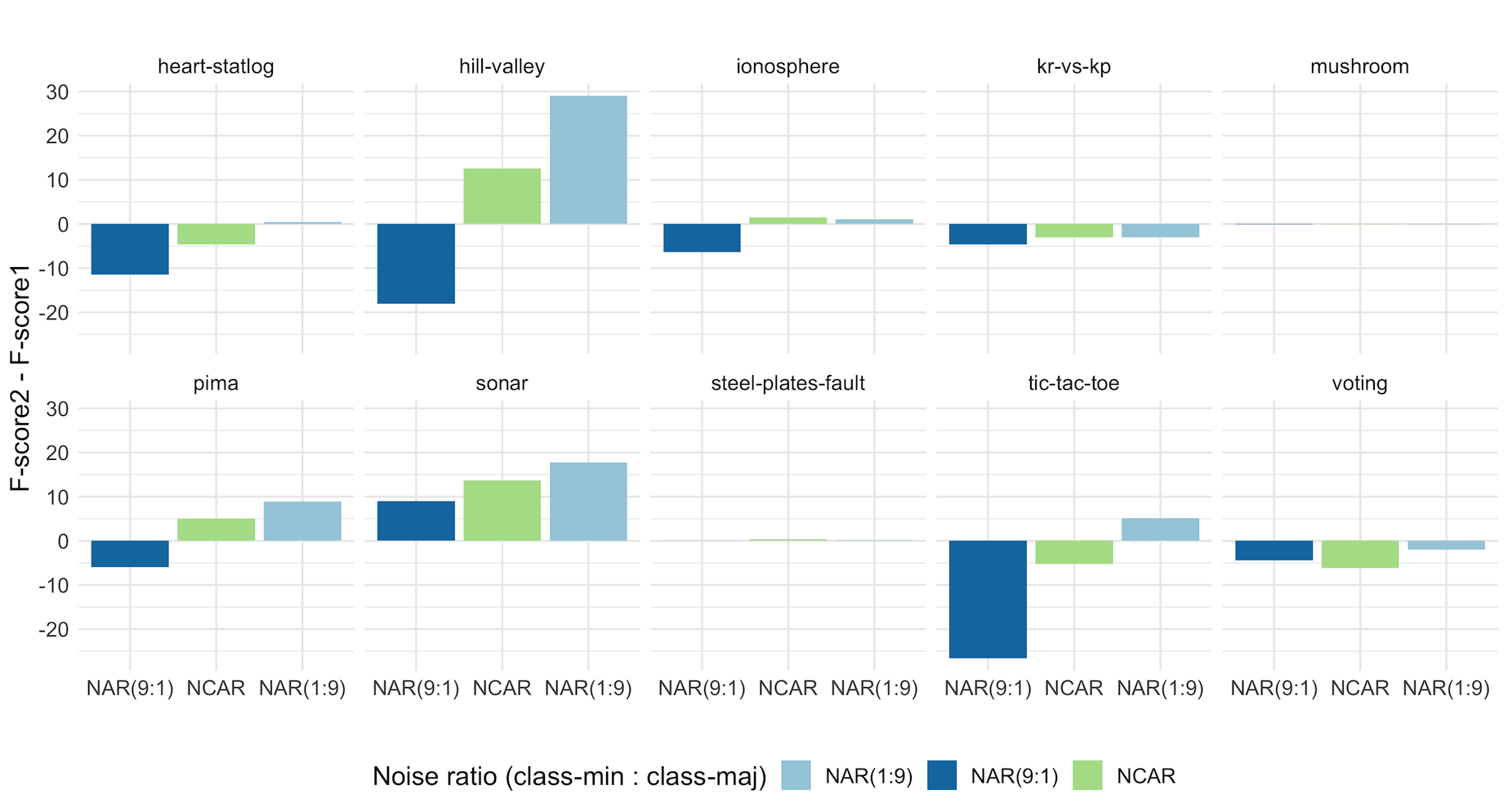}
 \caption{Variation in \textit{F-score} performance when IR increases from 50:50 (F-score1) to 20:80 (F-score2) in presence of a noise level of 15\%.}
 \label{fig:results_real_fscore_variation_15_2}
\end{figure}

Noise detection was impacted by class imbalance under the NAR model as expected.  For a better visualization of this result, Figures~\ref{fig:results_real_fscore_variation_15_1} and \ref{fig:results_real_fscore_variation_15_2} show the general behavior of noise detection under the NAR model when IR increases.  This can be observed by the more prominent and negative bar on the left side of the graphs in contrast to a smaller and positive bar on the right. This comportment was observed in all datasets, except for the \textit{sonar} dataset. This may imply that the noise model characteristics and class imbalance ratio, in general, have more influence on noise detection than the nature of the classification problem for the majority of problems.


\subsection{Noise Detection vs Ensemble Vote Thresholds}
\label{sec:real_different_thresholds}

\begin{figure}[!]
  \centering
  \includegraphics[width=5.5in]{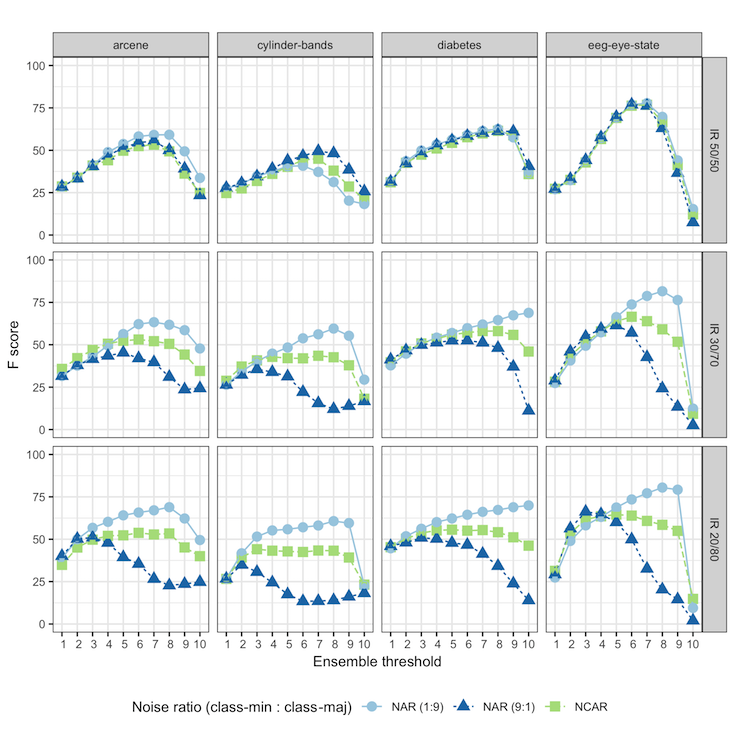}
 \caption{\textit{F-score} on \emph{arcene}, \emph{cylinder-bands}, \emph{diabetes}, and \emph{eeg-eye-state} datasets under different ensemble vote thresholds (where 1 = 10\%, 2 = 20\%,..,10 = 100\%) in presence of 15\% of noise.}
 \label{fig:results_real_threshold_1}
\end{figure}

\begin{figure}[!]
  \centering
  \includegraphics[width=5.5in]{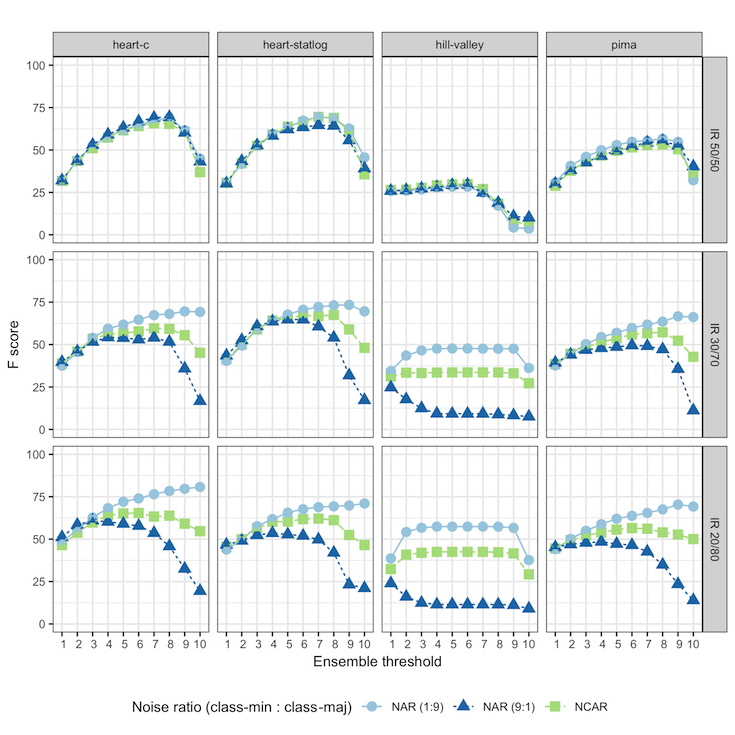}
 \caption{\textit{F-score} on \emph{heart-c}, \emph{heart-statlog}, \emph{hill-valley}, and \emph{pima} datasets under different ensemble vote thresholds (where 1 = 10\%, 2 = 20\%,..,10 = 100\%) in presence of 15\% of noise}
 \label{fig:results_real_threshold_2}
\end{figure}

Figures~\ref{fig:results_real_threshold_1} and \ref{fig:results_real_threshold_2} show the noise detection performance under different ensemble vote thresholds for each dataset at 15\% of noise level for the NAR and NCAR models.


As can be seen, for most datasets, similar behavior was observed: better noise detection was achieved with smaller threshold values under the NAR (9:1) model (i.e., more noise instances in the minority class than in the majority's), and higher threshold values under the NAR (1:9) model when IR is increased. Under the NCAR model, threshold values close to $L=5$ (majority vote) returned higher \textit{F-score} results. 

The above behavior is verified in a more or less pronounced way depending on the dataset. For example, in Figure~\ref{fig:results_real_threshold_1} for \textit{arcene} dataset, the best threshold under the NAR(9:1) model is $L = 7$ when IR is 50:50, $L = 5$ when IR is 30:70, and $L = 3$ for an IR of 20:80. Under the same settings, for \textit{pima} dataset in Figure~\ref{fig:results_real_threshold_2}, the best threshold under the NAR(9:1) model is $L = 8$ when IR is 50:50, $L = 6$  when IR is 30:70 and $L = 4$ for an IR of 20:80. The optimal values for each dataset are different, but the general behavior is the same. When varying the ensemble threshold, experiments show that a smaller number of voter algorithms delivers better noise detection under NAR(9:1) model, and that a greater threshold produces better performance under NAR(1:9). Finally, under NCAR models, the majority vote performs better.



Our results imply a change in the common practice adopted in the literature. The majority vote detection, widely used in related studies, corresponds to applying a default decision threshold. This is not the best option if it is expected a different noise level per class. Better thresholds can be set to improve noise detection performance. Different aspects like the class imbalance ratio and the noise model have to be considered.

\subsection{Statistical Tests}
\label{sec:stat_tests}

The Friedman test \citep{friedman1979}  was performed in order to compare the impact of all three noise models over the 228 problems (19 datasets\footnote{Mushroom dataset was removed because of its 100\% precision.} $\times$ 3 IR's $\times$ 4 different percentages of noise). The level of significance was set to $\alpha = 0.05$, i.e., 95\% confidence. 

The Friedman test shows a significant difference in the detection noise for the three models in certain contexts. As shown in Table~\ref{tab:results_friedmanRealSummary}, from the 152 data imbalanced problems analyzed, 77.63\% (118/152) presented a significant difference in the detection results. When considering only the problems with 20:80 IR, this number is equal to 88.16\%. On the other hand, when it comes to balanced datasets (76 of cases), only  18.42\% are significantly different. These results are aligned with the hypothesis discussed in the previous section. The choice of a noise generation model is more likely to impact detection results in data-imbalance problems.  

The influence of the amount of noise on ensemble detection was also tested. From the 57 problems for each percentage of noise, approximately half of the cases presented a significant difference (56.1\% for 5\% of noise, 	54.4\% for 10\%, 	63.2\% for 15\% and	57.9\% for 20\%). In this way, the amount of noise in data seems not to be as relevant as the IR on noise detection under different noise models. 

A second statistical analysis was also conducted in a pairwise fashion in order to verify if the noise models significantly improved/harmed the noise detection under certain contexts. In order to do so, the Wilcoxon non-parametric signed-rank test with the level of significance $\alpha = 0.05$ was used in all problems.

The tests were performed for each percentage of noise. As the results were equivalent (independently of the amount of noise), the following discussion will be regarding the noise percentage of 15\% shown in Table~\ref{tab:results_wilcoxonReal15}. 

Table~\ref{tab:results_wilcoxonReal15} presents a pairwise comparison of noise detection results for each noise model. The W/T/L denote the wins (better performance on noise detection), ties (equivalent performance on noise detection), and losses (worse performance on noise detection) produced by the noise models on the columns in comparison to the ones on the rows. For instance, the ensemble detector on problems under NAR(1:9) with 30:70 IR (column) performed 4 times better and 15 times worse (4/0/15) in comparison to the detection on problems under NAR(1:9) with 50:50 IR (row). For this same example, the $\textit{p-value} = 0.007$ implies there is a significant difference in the results.

\begin{table}[!]
\setlength\extrarowheight{2pt}
\begin{center}
\caption{Summary of Friedman test results on each problem.\label{tab:results_friedmanRealSummary}}
\begin{tabularx}{1\textwidth}{l|rr|rr|rr|rr|cc}
\toprule
\multirow{2}{*}{IR} & \multicolumn{8}{c|}{\textbf{Cases with significant difference}} & \multicolumn{2}{c}{\multirow{2}{*}{\textbf{Total per IR}}} \\ 
\cline{2-9}
& \multicolumn{2}{c|}{\textbf{5\% of noise}} & \multicolumn{2}{c|}{\textbf{10\% of noise}} & \multicolumn{2}{c|}{\textbf{15\% of noise}} & \multicolumn{2}{c|}{\textbf{20\% of noise}} & \multicolumn{2}{c}{}                                \\ \hline
\textbf{50:50}	& 6/19 & 	31.6\%	& 1/19 & 	5.3\%	& 6/19 & 	31.6\%	& 1/19 & 	5.3\%	& 14/76 & 	18.4\%	\\	
\textbf{30:70}	& 11/19 & 	57.9\%	& 14/19 & 	73.7\%	& 12/19 & 	63.2\%	& 14/19 & 	73.7\%	& 51/76 & 	67.1\%	\\	
\textbf{20:80}	& 15/19 & 	78.9\%	& 16/19 & 	84.2\%	& 18/19 & 	94.7\%	& 18/19 & 	94.7\%	& 67/76 & 	88.2\%	 \\	\cline{1-11}

\textbf{Total}	& 32/57 & 	56.1\%	& 31/57 & 	54.4\%	& 36/57 & 	63.2\%	& 33/57 & 	57.9\%	&	\\			
\end{tabularx}

\end{center}
\end{table}

Focusing on problems under the same IR, only one case showed a significant difference, as discussed previously. For balanced data, the NAR(1:9) model (more noise in the majority class) produced a positive impact on noise detection, performing 15 times out of 19 better than the detection under NCAR, although with a $\textit{p-value} = 0.038$. For the case of 30:70 IR, the NAR(9:1) model (more noise in the minority class) harmed detection as its performance was worse 17 times when compared to the detection under NCAR and NAR(1:9) models with a really small $\textit{p-value}$. This was also verified when data was exposed to a different percentage of noise. Lastly, when increasing IR (20:80), tests showed noise detection is significantly improved under NAR(1:9) in comparison to the other noise models as the ensemble detector performed better in all problems (19/0/0).

\addtocounter{table}{-1} 
\begin{table}[t]
\setlength\extrarowheight{2pt}
\begin{center}

\caption{Wilcoxon test on real problems when there is 15\% noise in data. W \textbackslash T \textbackslash L = wins\textbackslash ties\textbackslash losses. p-value $< 0.05$ are highlighted.\label{tab:results_wilcoxonReal15}}

\begin{footnotesize}
\begin{tabular}{lll|ll|lll|lll}														
\toprule														
\multirow{2}{*}{\textbf{IR}}    & \multicolumn{2}{c|}{\multirow{2}{*}{\textbf{Noise Model}}} & \multicolumn{2}{c|}{\textbf{50:50}}    & \multicolumn{3}{c|}{\textbf{30:70}}                    & \multicolumn{3}{c}{\textbf{20:80}}                    \\														
& \multicolumn{2}{l|}{}                                       & \textbf{NCAR} & \textbf{\begin{tabular}[c]{@{}c@{}}NAR\\ (1:9)\end{tabular}} & \textbf{\begin{tabular}[c]{@{}c@{}}NAR\\ (9:1)\end{tabular}} & \textbf{NCAR} & \textbf{\begin{tabular}[c]{@{}c@{}}NAR\\ (1:9)\end{tabular}} & \textbf{\begin{tabular}[c]{@{}c@{}}NAR\\ (9:1)\end{tabular}} & \textbf{NCAR} & \textbf{\begin{tabular}[c]{@{}c@{}}NAR\\ (1:9)\end{tabular}} \\ \hline														
\multirow{6}{*}{\textbf{50:50}}	&	\multirow{2}{*}{\textbf{NAR(9:1)}}	&	\textbf{W/T/L}	& 5/0/14	& 11/1/7	& 16/0/3	& 17/0/2	& 15/0/4	& 17/0/2	& 17/0/2	& 16/0/3	\\	
	&		&	\textbf{p-value}	& 0.073	& 0.360	& \textbf{0.001}	& \textbf{0.000}	& \textbf{0.001}	& \textbf{0.001}	& \textbf{0.001}	& \textbf{0.001}	\\	
	&	\multirow{2}{*}{\textbf{NCAR}}	&	\textbf{W/T/L}	& 	& 15/0/4	& 11/0/8	& 10/0/9	& 9/0/10	& 13/0/6	& 8/0/11	& 12/0/7	\\	
	&		&	\textbf{p-value}	& 	& \textbf{0.038}	& 0.481	& 0.952	& 0.324	& 0.409	& 0.952	& 0.153	\\	
	&	\multirow{2}{*}{\textbf{NAR(1:9)}}	&	\textbf{W/T/L}	&	& 	& 6/0/13	& 4/0/15	& 4/0/15	& 3/0/16	& 3/0/16	& 3/0/16	\\	
	&		&	\textbf{p-value}	&	& 	& \textbf{0.021}	& \textbf{0.005}	& \textbf{0.007}	& \textbf{0.003}	& \textbf{0.002}	& \textbf{0.004}	\\	\hline
\multirow{6}{*}{\textbf{30:70}}	&	\multirow{2}{*}{\textbf{NAR(9:1)}}	&	\textbf{W/T/L}	&	&	& 	& 17/0/2	& 17/0/2	& 16/0/3	& 16/0/3	& 17/0/2	\\	
	&		&	\textbf{p-value}	&	&	& 	& \textbf{0.000}	& \textbf{0.000}	& \textbf{0.012}	& \textbf{0.001}	& \textbf{0.000}	\\	
	&	\multirow{2}{*}{\textbf{NCAR}}	&	\textbf{W/T/L}	&	&	&	& 	& 17/0/2	& 4/0/15	& 9/0/10	& 15/0/4	\\	
	&		&	\textbf{p-value}	&	&	&	& 	& \textbf{0.000}	& \textbf{0.004}	& 0.856	& \textbf{0.004}	\\	
	&	\multirow{2}{*}{\textbf{NAR(1:9)}}	&	\textbf{W/T/L}	&	&	&	&	& 	& 1/0/18	& 4/0/15	& 7/0/12	\\	
	&		&	\textbf{p-value}	&	&	&	&	& 	& \textbf{0.000}	& \textbf{0.001}	& \textbf{0.035}	\\	\hline
\multirow{6}{*}{\textbf{20:80}}	&	\multirow{2}{*}{\textbf{NAR(9:1)}}	&	\textbf{W/T/L}	&	&	&	&	&	& 	& 18/0/1	& 19/0/0	\\	
	&		&	\textbf{p-value}	&	&	&	&	&	& 	& \textbf{0.000}	& \textbf{0.000}	\\	
	&	\multirow{2}{*}{\textbf{NCAR}}	&	\textbf{W/T/L}	&	&	&	&	&	&	& 	& 19/0/0	\\	
	&		&	\textbf{p-value}	&	&	&	&	&	&	& 	& \textbf{0.000}	\\	
\bottomrule														
\end{tabular}														
\end{footnotesize}

\end{center}
\end{table}


\section{Conclusions}
\label{sec:conclusion}

Many studies have focused their attention on data quality issues due to its importance in {ML} applications and the known fact that real-world datasets frequently contain noise~\citep{survey}. 

Noise can be presented in data in its attributes and also in its classes \citep{Zhu-attr}. This work focused the studies on class noise (also label noise). For this type of problem, the \textit{classification noise filtering} approach is usually applied to remove data irregularities before the learning step. The most common filtering consists of using the predictions of an ensemble of algorithms so that instances are removed upon a wrong classification  \citep{Brodley}\citep{Sluban-diversity}\citep{unlabled2018}.

To evaluate Noise Filters, simulated noise is usually injected into a dataset, and then analyses can be performed on the results \citep{GARCIA2019}. In \cite{survey}, three different label noise generation for injecting noise are presented: (1) {NCAR}, in which the probability of an instance being noisy is random, (2) {NAR}, the probability of an instance being noisy depends on its label, and (3) {NNAR}, the probability of an instance being noisy also depends on its attributes.

Although there are many approaches to model different noise behaviors, in many previous works \citep{Sluban-diversity} \citep{Brodley} \citep{saesSMOTE} \citep{GARCIA2019}, one type of noise is chosen over another without considering the different impacts of each. Also, despite the majority and consensus vote approaches being the most common ensemble voting schemes used for filtering noise, studies   \citep{threshold2005}\citep{threshold2018} have shown that selecting adequate values for the ensemble threshold can lead to superior results.

This work presented an empirical study focused on ensemble-based noise detection and its performance under three different noise models generation: NCAR, where noise is equally distributed among class,  NAR model by applying more noise in the majority class, and NAR model by employing more noise in the minority classes. The relation between detection performance and injected noise model was assessed through performance measures ({\it F-score}, Precision, Recall), considering different ratios of inserted noise and imbalance class configurations. The impact produced on filtering performance was also evaluated under different ensemble thresholds. 

As conclusions, we could observe that noise detection was not affected by the noise model in balanced data. In contrast, when dealing with imbalanced data, we faced two different scenarios for noise detection under the NAR model. Firstly, the addition of noise in the minority class harmed the noise detection in all problems. Moreover, by applying more noise in the majority class, noise detection improved in most cases. Besides, increasing the IR from 30:70 to 20:80 resulted in an even worse noise detection when the minority class was the noisiest one, and the opposite was verified when the majority class had more noise.

The experiments performed with different ensemble thresholds showed that not always \textit{majority} and \textit{consensus} voting are the best options. Better noise detection was achieved for both models NCAR and NAR(9:1) if less than 50\% of the algorithms were selected. On the other hand, better noise detection was achieved for NAR(1:9) if more than 50\% of the algorithms were selected.

As future work, we intend to investigate the NNAR model and evaluate other noise filtering techniques, expanding the work to multi-class problems as well.

\section*{Acknowledgment}
This research has been supported in part by the following Brazilian agencies: CAPES, CNPq and FACEPE.

\bibliographystyle{unsrtnat}
\bibliography{references}  

\begin{thebibliography}{32}
\providecommand{\natexlab}[1]{#1}
\providecommand{\url}[1]{\texttt{#1}}
\expandafter\ifx\csname urlstyle\endcsname\relax
  \providecommand{\doi}[1]{doi: #1}\else
  \providecommand{\doi}{doi: \begingroup \urlstyle{rm}\Url}\fi

\bibitem[Han et~al.(2012)Han, Kamber, and Pei]{Han}
Jiawei Han, Micheline Kamber, and Jian Pei.
\newblock \emph{Data Mining: Concepts and Techniques}.
\newblock Morgan Kaufmann Publishers, 2012.

\bibitem[Zhu and Wu(2004)]{Zhu-attr}
X.~Zhu and X.~Wu.
\newblock Class noise vs. attribute noise: A quantitative study.
\newblock In \emph{Artifficial Intelligence Review}, pages 177--210, 2004.
\newblock \doi{10.1007/s10462-004-0751-8}.

\bibitem[Frenay and Verleysen(2014)]{survey}
B.~Frenay and M.~Verleysen.
\newblock Classification in the presence of label noise: A survey.
\newblock \emph{IEEE Transactions on Neural Networks and Learning Systems},
  25\penalty0 (5):\penalty0 845--869, 2014.
\newblock \doi{10.1109/TNNLS.2013.2292894}.

\bibitem[Brodley and Friedl(1999)]{Brodley}
C.~E. Brodley and M.~A. Friedl.
\newblock Identifying mislabeled training data.
\newblock In \emph{Artifficial Intelligence Review}, page 131–167, 1999.
\newblock \doi{10.1613/jair.606}.

\bibitem[Sluban and Lavrač(2015)]{Sluban-diversity}
Borut Sluban and Nada Lavrač.
\newblock Relating ensemble diversity and performance: A study in class noise
  detection.
\newblock \emph{Neurocomputing}, 160:\penalty0 120--131, 2015.
\newblock ISSN 0925-2312.
\newblock \doi{https://doi.org/10.1016/j.neucom.2014.10.086}.
\newblock URL
  \url{https://www.sciencedirect.com/science/article/pii/S0925231215001265}.

\bibitem[Guan et~al.(2018)Guan, Wei, Yuan, Han, Tian, Al-Dhelaan, and
  Al-Dhelaan]{unlabled2018}
D.~Guan, H.~Wei, W.~Yuan, G.~Han, Y.~Tian, M.~Al-Dhelaan, and A.~Al-Dhelaan.
\newblock Improving label noise filtering by exploiting unlabeled data.
\newblock \emph{IEEE Access}, 6:\penalty0 11154--11165, 2018.
\newblock \doi{10.1109/ACCESS.2018.2807779}.

\bibitem[Garcia et~al.(2019)Garcia, Lehmann, {de Carvalho}, and
  Lorena]{GARCIA2019}
Luís~P.F. Garcia, Jens Lehmann, André~C.P.L.F. {de Carvalho}, and Ana~C.
  Lorena.
\newblock New label noise injection methods for the evaluation of noise
  filters.
\newblock \emph{Knowledge-Based Systems}, 163:\penalty0 693--704, 2019.
\newblock ISSN 0950-7051.
\newblock \doi{https://doi.org/10.1016/j.knosys.2018.09.031}.
\newblock URL
  \url{https://www.sciencedirect.com/science/article/pii/S0950705118304829}.

\bibitem[Sáez et~al.(2015)Sáez, Luengo, Stefanowski, and Herrera]{saesSMOTE}
José~A. Sáez, Julián Luengo, Jerzy Stefanowski, and Francisco Herrera.
\newblock Smote–ipf: Addressing the noisy and borderline examples problem in
  imbalanced classification by a re-sampling method with filtering.
\newblock \emph{Information Sciences}, 291:\penalty0 184--203, 2015.
\newblock ISSN 0020-0255.
\newblock \doi{https://doi.org/10.1016/j.ins.2014.08.051}.
\newblock URL
  \url{https://www.sciencedirect.com/science/article/pii/S0020025514008561}.

\bibitem[B.~Sluban and Lavra(2015)]{Sluban-advances}
D.~Gamberger B.~Sluban and N.~Lavra.
\newblock Advances in class noise detection.
\newblock In \emph{European Conference on Artificial Intelligence}, pages
  1105--1106, 2015.
\newblock \doi{10.3233/978-1-60750-606-5-1105}.

\bibitem[Lorena and Carvalho(2004)]{LORENA2004}
Ana~C. Lorena and Andre C. P. L. F.~de Carvalho.
\newblock {Evaluation of noise reduction techniques in the splice junction
  recognition problem}.
\newblock \emph{{Genetics and Molecular Biology}}, 27:\penalty0 665 -- 672, 00
  2004.
\newblock ISSN 1415-4757.
\newblock \doi{10.1590/S1415-47572004000400031}.

\bibitem[Abell{\'a}n and Masegosa(2010)]{abellan-baggingDT}
Joaqu{\'i}n Abell{\'a}n and Andr{\'e}s~R. Masegosa.
\newblock Bagging decision trees on data sets with classification noise.
\newblock In Sebastian Link and Henri Prade, editors, \emph{Foundations of
  Information and Knowledge Systems}, pages 248--265, Berlin, Heidelberg, 2010.
  Springer Berlin Heidelberg.
\newblock \doi{10.1007/978-3-642-11829-6_17}.

\bibitem[Bootkrajang(2016)]{B2016star}
Jakramate Bootkrajang.
\newblock A generalised label noise model for classification in the presence of
  annotation errors.
\newblock \emph{Neurocomputing}, 192:\penalty0 61--71, 2016.
\newblock ISSN 0925-2312.
\newblock \doi{https://doi.org/10.1016/j.neucom.2015.12.106}.
\newblock URL
  \url{https://www.sciencedirect.com/science/article/pii/S0925231216002551}.
\newblock Advances in artificial neural networks, machine learning and
  computational intelligence.

\bibitem[Lawrence and Scholkopf(2001)]{FisherLawrence:2001}
Neil~D. Lawrence and Bernhard Scholkopf.
\newblock Estimating a kernel fisher discriminant in the presence of label
  noise.
\newblock In \emph{Proceedings of the Eighteenth International Conference on
  Machine Learning}, pages 306--313, San Francisco, CA, USA, 2001. Morgan
  Kaufmann Publishers Inc.
\newblock ISBN 1-55860-778-1.

\bibitem[Bootkrajang and Kabán(2014)]{BOOTKRAJANG2014}
Jakramate Bootkrajang and Ata Kabán.
\newblock Learning kernel logistic regression in the presence of class label
  noise.
\newblock \emph{Pattern Recognition}, 47\penalty0 (11):\penalty0 3641--3655,
  2014.
\newblock ISSN 0031-3203.
\newblock \doi{https://doi.org/10.1016/j.patcog.2014.05.007}.
\newblock URL
  \url{https://www.sciencedirect.com/science/article/pii/S0031320314001927}.

\bibitem[Biggio et~al.(2011)Biggio, Nelson, and Laskov]{biggio2011}
Battista Biggio, Blaine Nelson, and Pavel Laskov.
\newblock Support vector machines under adversarial label noise.
\newblock In Chun-Nan Hsu and Wee~Sun Lee, editors, \emph{Proceedings of the
  Asian Conference on Machine Learning}, volume~20 of \emph{Proceedings of
  Machine Learning Research}, pages 97--112, South Garden Hotels and Resorts,
  Taoyuan, Taiwain, 14--15 Nov 2011. PMLR.

\bibitem[Sun et~al.(2007)Sun, Zhao, Wang, and Chen]{sun-adHocMeasure}
J.~Sun, F.~Zhao, C.~Wang, and S.~Chen.
\newblock Identifying and correcting mislabeled training instances.
\newblock In \emph{Future Generation Communication and Networking (FGCN 2007)},
  volume~1, pages 244--250, 2007.
\newblock \doi{10.1109/FGCN.2007.146}.

\bibitem[Smith et~al.(2014)Smith, Martinez, and Giraud-Carrier]{Smith2014}
Michael~R. Smith, Tony Martinez, and Christophe Giraud-Carrier.
\newblock An instance level analysis of data complexity.
\newblock \emph{Machine Learning}, 95\penalty0 (2):\penalty0 225--256, May
  2014.
\newblock \doi{10.1007/s10994-013-5422-z}.

\bibitem[Zhu et~al.(2003)Zhu, Wu, and Chen]{zhu-large}
X.~Zhu, X.~Wu, and Q.~Chen.
\newblock Eliminating class noise in large datasets.
\newblock In \emph{In 20th International Conference on Machine Learning
  (ICML)}, page 920–927, 2003.

\bibitem[García-Gil et~al.(2019)García-Gil, Luengo, García, and
  Herrera]{GARCIAGIL2019bigData}
Diego García-Gil, Julián Luengo, Salvador García, and Francisco Herrera.
\newblock Enabling smart data: Noise filtering in big data classification.
\newblock \emph{Information Sciences}, 479:\penalty0 135--152, 2019.
\newblock ISSN 0020-0255.
\newblock \doi{https://doi.org/10.1016/j.ins.2018.12.002}.
\newblock URL
  \url{https://www.sciencedirect.com/science/article/pii/S0020025518309460}.

\bibitem[Malossini et~al.(2006)Malossini, Blanzieri, and Ng]{LOOPC}
Andrea Malossini, Enrico Blanzieri, and Raymond~T. Ng.
\newblock {Detecting potential labeling errors in microarrays by data
  perturbation}.
\newblock \emph{Bioinformatics}, 22\penalty0 (17):\penalty0 2114--2121, 06
  2006.
\newblock ISSN 1367-4803.
\newblock \doi{10.1093/bioinformatics/btl346}.
\newblock URL \url{https://doi.org/10.1093/bioinformatics/btl346}.

\bibitem[Wilson and Martinez(2000)]{Wilson2000}
D.~Randall Wilson and Tony~R. Martinez.
\newblock Reduction techniques for instance-based learning algorithms.
\newblock \emph{Machine Learning}, 38\penalty0 (3):\penalty0 257--286, Mar
  2000.
\newblock \doi{10.1023/A:1007626913721}.

\bibitem[Kanj et~al.(2016)Kanj, Abdallah, Den{\oe}ux, and Tout]{Kanj2016ENN}
Sawsan Kanj, Fahed Abdallah, Thierry Den{\oe}ux, and Kifah Tout.
\newblock Editing training data for multi-label classification with the
  k-nearest neighbor rule.
\newblock \emph{Pattern Analysis and Applications}, 19\penalty0 (1):\penalty0
  145--161, Feb 2016.
\newblock \doi{10.1007/s10044-015-0452-8}.

\bibitem[Yuan et~al.(2018)Yuan, Guan, Ma, and Khattak]{YUAN2018psma}
Weiwei Yuan, Donghai Guan, Tinghuai Ma, and Asad~Masood Khattak.
\newblock Classification with class noises through probabilistic sampling.
\newblock \emph{Inf. Fusion}, 41\penalty0 (C):\penalty0 57–67, may 2018.
\newblock ISSN 1566-2535.
\newblock \doi{10.1016/j.inffus.2017.08.007}.
\newblock URL \url{https://doi.org/10.1016/j.inffus.2017.08.007}.

\bibitem[Verbaeten and Van~Assche(2003)]{comitee}
Sofie Verbaeten and Anneleen Van~Assche.
\newblock Ensemble methods for noise elimination in classification problems.
\newblock In Terry Windeatt and Fabio Roli, editors, \emph{Multiple Classifier
  Systems}, pages 317--325, Berlin, Heidelberg, 2003. Springer Berlin
  Heidelberg.
\newblock ISBN 978-3-540-44938-6.
\newblock \doi{10.1007/3-540-44938-8_32}.

\bibitem[Sluban et~al.(2014)Sluban, Gamberger, and Lavra{\v{c}}]{Sluban2014}
Borut Sluban, Dragan Gamberger, and Nada Lavra{\v{c}}.
\newblock Ensemble-based noise detection: noise ranking and visual performance
  evaluation.
\newblock \emph{Data Mining and Knowledge Discovery}, 28\penalty0 (2):\penalty0
  265--303, Mar 2014.
\newblock \doi{10.1007/s10618-012-0299-1}.

\bibitem[Khoshgoftaar et~al.(2005)Khoshgoftaar, Zhong, and
  Joshi]{threshold2005}
Taghi~M. Khoshgoftaar, Shi Zhong, and Vedang Joshi.
\newblock Enhancing software quality estimation using ensemble-classifier based
  noise filtering.
\newblock \emph{Intell. Data Anal.}, 9\penalty0 (1):\penalty0 3--27, January
  2005.
\newblock ISSN 1088-467X.
\newblock \doi{10.3233/IDA-2005-9102}.

\bibitem[Sabzevari et~al.(2018)Sabzevari, Martnez-Muoz, and
  Surez]{threshold2018}
Maryam Sabzevari, Gonzalo Martnez-Muoz, and Alberto Surez.
\newblock A two-stage ensemble method for the detection of class-label noise.
\newblock \emph{Neurocomput.}, 275\penalty0 (C):\penalty0 2374–2383, jan
  2018.
\newblock ISSN 0925-2312.
\newblock \doi{10.1016/j.neucom.2017.11.012}.
\newblock URL \url{https://doi.org/10.1016/j.neucom.2017.11.012}.

\bibitem[Gilbert et~al.(2016)Gilbert, Martin, Donovan, Lane, Hamdy, Neal, and
  Metcalfe]{differencial}
Rebecca Gilbert, Richard~M Martin, Jenny Donovan, J~Athene Lane, Freddie Hamdy,
  David~E Neal, and Chris Metcalfe.
\newblock Misclassification of outcome in case–control studies: Methods for
  sensitivity analysis.
\newblock \emph{Statistical Methods in Medical Research}, 25\penalty0
  (5):\penalty0 2377--2393, 2016.
\newblock \doi{10.1177/0962280214523192}.
\newblock URL \url{https://doi.org/10.1177/0962280214523192}.
\newblock PMID: 25217446.

\bibitem[AlcalA-Fdez et~al.(2011)AlcalA-Fdez, Fernandez, Luengo, Derrac,
  GarcIa, Sanchez, and Herrera]{KEEL}
J.~AlcalA-Fdez, A.~Fernandez, J.~Luengo, J.~Derrac, S.~GarcIa, L.~Sanchez, and
  F.~Herrera.
\newblock Keel data-mining software tool: Data set repository, integration of
  algorithms and experimental analysis framework.
\newblock \emph{Journal of Multiple-Valued Logic and Soft Computing},
  17\penalty0 (2-3):\penalty0 255--287, 2011.

\bibitem[Dua and Karra~Taniskidou(2017)]{UCI:2017}
Dheeru Dua and Efi Karra~Taniskidou.
\newblock {UCI} machine learning repository, 2017.
\newblock URL \url{http://archive.ics.uci.edu/ml}.

\bibitem[Vanschoren et~al.(2014)Vanschoren, van Rijn, Bischl, and
  Torgo]{OpenML2013}
Joaquin Vanschoren, Jan~N. van Rijn, Bernd Bischl, and Luis Torgo.
\newblock Openml: Networked science in machine learning.
\newblock \emph{SIGKDD Explor. Newsl.}, 15\penalty0 (2):\penalty0 49–60, jun
  2014.
\newblock ISSN 1931-0145.
\newblock \doi{10.1145/2641190.2641198}.
\newblock URL \url{https://doi.org/10.1145/2641190.2641198}.

\bibitem[Friedman and Rafsky(1979)]{friedman1979}
Jerome~H. Friedman and Lawrence~C. Rafsky.
\newblock {Multivariate Generalizations of the Wald-Wolfowitz and Smirnov
  Two-Sample Tests}.
\newblock \emph{The Annals of Statistics}, 7\penalty0 (4):\penalty0 697 -- 717,
  1979.
\newblock \doi{10.1214/aos/1176344722}.
\newblock URL \url{https://doi.org/10.1214/aos/1176344722}.

\end{thebibliography}

\end{document}